%% file: iclr2026_conference.tex
\documentclass{article} 
\usepackage{iclr2026_arxiv,times}
\iclrfinalcopy

\input{math_commands.tex}

\usepackage{hyperref}
\usepackage{url}

\usepackage[utf8]{inputenc} 
\usepackage[T1]{fontenc}    
\usepackage{hyperref}       
\usepackage{url}            
\usepackage{booktabs}       
\usepackage{amsfonts}       
\usepackage{nicefrac}       
\usepackage{microtype}      
\usepackage[table, dvipsnames]{xcolor}
\usepackage{xcolor}         

\usepackage{times}
\usepackage{latexsym}
\usepackage{amsmath}
\usepackage{amsfonts}
\usepackage{graphicx}
\usepackage{amssymb}
\usepackage{pifont}
\newcommand{\cmark}{\ding{51}}%
\newcommand{\xmark}{\ding{55}}%
\usepackage{hyperref}
\usepackage[nameinlink]{cleveref}
\usepackage{soul}
\usepackage{algorithm}
\usepackage{algpseudocode}
\usepackage{booktabs}
\usepackage{multirow}
\usepackage{arydshln}
\usepackage{adjustbox}
\usepackage{lipsum}
\usepackage{xcolor}
\usepackage{wrapfig}
\usepackage{subcaption}
\usepackage{graphicx}

\newcommand{\react}[0]{\textsc{React}}

\newcommand{\internalDyna}[0]{{Dyna-Think}}
\newcommand{\internalDynatraining}[0]{{Dyna-Think Dyna Training}}
\newcommand{\internalDynatrainingshort}[0]{{DDT}}

\newcommand{\roneminimal}[0]{$\pi_{\mathcal{W}}(\theta)$}
\newcommand{\policy}[0]{$\pi(\theta)$}

\newcommand{\worldmodelmu}[0]{$\mathcal{W}(\mu)$}

\newcommand{\directdistillshort}[0]{\textsc{Direct Distill}}
\newcommand{\wmre}[0]{{Dyna-Think Imitation Learning}}
\newcommand{\wmreshort}[0]{\textsc{DIT}}

\newcommand{\rft}[0]{\textsc{RFT}}
\newcommand{\rftlong}[0]{Reinforcement Finetuning}
\newcommand{\judgeprefinetuneshort}[0]{\internalDynatrainingshort{}($\hat{\mathcal{T}}_{\textrm{critic}}$)}
\newcommand{\wmdiffprefinetuneshort}[0]{\internalDynatrainingshort{}($\hat{\mathcal{T}}_{\Delta}$)}
\newcommand{\wmprefinetuneshort}[0]{\internalDynatrainingshort{}($\hat{\mathcal{T}}$)}

\definecolor{light-gray}{gray}{0.9}
\newcommand{\ie}{i.e.,\ }
\newcommand{\eg}{e.g.,\ }

\title{
    \textsc{\internalDyna{}}: Synergizing Reasoning, Acting, and World Model Simulation in AI Agents
}

\author{Xiao Yu$^1$, \, Baolin Peng$^{2\dagger}$, \, Ruize Xu$^1$, \, Michel Galley$^2$, \, Hao Cheng$^2$, \\ \bf{Suman Nath$^{2}$,} \, \bf{Jianfeng Gao$^{2*}$ \& Zhou Yu$^{13}$}\thanks{Equal Advisory Contribution; $^\dagger$ Project Lead} \\
$^1$Columbia University, NY \, $^2$Microsoft Research, Redmond\,\, $^3$Arklex.ai, NY \\
\texttt{\{xy2437, zy2461\}@columbia.edu} \\
\texttt{\{baolinpeng, jfgao\}@microsoft.com}
}

\begin{document}

\maketitle

\begin{abstract}
Recent progress in reasoning with large language models (LLMs), such as DeepSeek-R1, demonstrates impressive capabilities in domains like mathematics and coding, by exhibiting complex cognitive behaviors such as verification, goal decomposition, and self-reflection.
However, it is unclear what behavior is effective and what behavior is missing for agentic tasks.
In this work, we propose \internalDyna{}, a thinking framework that integrates \textit{planning with an internal world model} with reasoning and acting to enhance AI agent performance.
To enable \internalDyna{}, we propose \wmre{} (\wmreshort{}) and \internalDynatraining{} (\internalDynatrainingshort{}).
To initialize a policy with \internalDyna{}, \wmreshort{} reconstructs the thinking process of R1 to focus on performing world model simulation relevant to the proposed (and planned) action, and trains the policy using this reconstructed data.
To enhance \internalDyna{}, \internalDynatrainingshort{} uses a two-stage training process to first improve the agent's world modeling ability for next state prediction and critique generation, and then improve the agent's action via policy training.
We evaluate our methods on OSWorld and WindowsAgentArena, and demonstrate that \internalDyna{} improves the agent's in-domain and out-of-domain performance, achieving similar best-of-n performance compared to R1 while generating 2x less tokens on average.
Our extensive empirical studies reveal that 1) using critique generation for world model training is effective to improving policy performance; and 2) improving a world model is effective to improving the performance of its AI agent.
Our results suggest a promising research direction to integrate world-model simulation into AI agents to enhance their reasoning, planning, and acting capabilities.
\end{abstract}

\section{Introduction}
Autonomous AI agents powered by large language models (LLMs) have offered substantial promise in real-world applications, automating digital tasks such as software engineering \citep{jimenez2024swebenchlanguagemodelsresolve,wang2024opendevinopenplatformai,yang2024sweagentagentcomputerinterfacesenable}, web navigation \citep{liu2018miniwobpp,yao2023webshopscalablerealworldweb,zhou2024webarenarealisticwebenvironment,koh2024vwa}, computer-use \citep{xie2024osworldbenchmarkingmultimodalagents,computer-use,qin2025uitarspioneeringautomatedgui}, and mobile device control \citep{rawles2023androidwildlargescaledataset,rawles2024androidworlddynamicbenchmarkingenvironment,trivedi2024appworldcontrollableworldapps}.
As many computer use tasks require agents to interact with complex environments to achieve long-term objectives, one critical challenge is to reason and act efficiently over a large decision space.


Recent methods of test-time scaling \citep{snell2024scalingllmtesttimecompute} offer a potentially promising solution.
For example, \citet{yu2023promptbasedmontecarlotreesearch,yu2024exactteachingaiagents,zhou2024languageagenttreesearch} show that LLM agents can significantly improve performance using search algorithms such as Monte Carlo Tree Search (MCTS) to allow additional interactions with the environment before decision-making.
However, these methods require a large amount of expensive, time-consuming interactions with a real (or a separately learned) world model, limiting their applicability in real-world scenarios. 
Alternatively, many recent work \citep{o1,valmeekam2024llmscantplanlrms,zhou2025largereasoningmodelsagent} finds that LLM agents can also improve their performance by ``thinking'' longer, effectively internalizing parts of search into their reasoning process.
For example, models such as OpenAI o1/o3 \citep{o1,o3}, Claude-3.7-Sonnet \citep{claude-3.7}, and DeepSeek-R1 \citep{deepseekai2025deepseekr1incentivizingreasoningcapability} generate significantly longer reasoning chains when faced with challenging tasks, exhibiting behaviors such as self-reflection, goal decomposition, verification, exploration, and more \citep{gandhi2025cognitivebehaviorsenableselfimproving}.
Although many of these behaviors are crucial for solving challenging long-horizon tasks, recent work also finds that these models suffer from problems such as overthinking and fact-ignoring \citep{cuadron2025dangeroverthinkingexaminingreasoningaction,zhou2025largereasoningmodelsagent}, and are less efficient in certain tasks compared to non-reasoning models such as GPT-4o \citep{zhou2025largereasoningmodelsagent}.
It is thus unclear \emph{what} type of ``thinking'' is crucial to an agent's performance for long-horizon tasks, and \emph{how} to improve it via learning from experiences.

\input{floats/dyna_think_example}

The successes in Go, games, and robotics reveal that modeling and interacting with the environment is key to solving challenging long-horizon tasks.
The Dyna algorithms \citep{dyna} offer a promising approach to combine real-world interactions with planning via simulation to improve an agent's policy.
However, modeling the whole environment for tasks such as computer-use (\eg predicting screen content after entering `amazon.com' in chrome) is challenging \citep{pathak2017curiositydrivenexplorationselfsupervisedprediction,wang2024languagemodelsservetextbased,fang2025webevolverenhancingwebagent,gu2025llmsecretlyworldmodel}.
In contrast, research in cognitive sciences \citep{human-vision,Rao99} shows that the human brain encodes a \emph{compressed} representation of the external world, capturing only statistical regularities and meaningful structures related to current tasks.
We thus propose \internalDyna{}, a thinking framework that performs \emph{``compressed'' world model simulation} based on model-generated states and critiques (rewards),
and \emph{integrates it with reasoning and acting} to improve an AI agent's performance. \Cref{fig:dyna_think_example} illustrates the difference between the Dyna-Think agent and DeepSeek-R1 using an example.

The \internalDyna{} framework consists of \wmre{} (\wmreshort{}) and \internalDynatraining{} (\internalDynatrainingshort{}).
\wmreshort{} improves a baseline agent policy by first distilling simulated experiences from R1-generated thinking tokens, and then revising the baseline policy using the simulated experiences via supervised learning.
We finds models after \wmreshort{} achieve a similar performance to R1-distilled models, but with 2X less thinking tokens on average.
To further improve \wmreshort{}, we propose \internalDynatrainingshort{}, is an extension to the Dyna-Q method \citep{Sutton1998} that combines policy and world model learning during online training.
Similar to Dyna-Q, \internalDynatrainingshort{} first constructs policy and world model training data using online rollouts in a real (digital) environment.
Different from Dyna-Q, \internalDynatrainingshort{} is applicable to the problems with arbitrarily large state-action space by leveraging the prior knowledge encoded in a pre-trained LLM, and performs both policy learning and world model training based on \emph{a single LLM}.
During world model training, the model encodes state transitions by optimizing training objectives of next-state prediction and critiquing the states generated by its own.
During policy learning, the agent's actions are improved using successful rollouts via reinforcement learning.

We evaluate \internalDyna{} on OSWorld \citep{xie2024osworldbenchmarkingmultimodalagents} and WindowsAgentArena \citep{bonatti2024windowsagentarenaevaluating}.
Results show that \internalDyna{} improves the agent's in-domain and out-of-domain performance compared to only performing policy training (\eg reinforcement learning) or training a separate world model (\eg as in Dyna-Q), and that
\internalDyna{} leads to highly capable and cost-effective agents. 
For example, our 32B-parameter \internalDyna{} model achieves a similar best-of-n performance compared to the much larger 685B-parameter R1 model, with 2x less tokens on average. 
Detailed analysis also 
reveals that 1) the critique-style world model training is effective in improving policy; and 
2) AI agents with a stronger world model achieves better performance.
Thus, our work demonstrates the potential of integrating planning and learning to develop future agents powered by reasoning models.
\section{Related Work}
\label{sec:Related Work}
\paragraph{Computer-Use Agents}
Computer-use agents powered by large (multimodal) language models aim to automate tasks by controlling a computer, typically via GUI interactions with a virtual Ubuntu/Windows \citep{xie2024osworldbenchmarkingmultimodalagents,bonatti2024windowsagentarenaevaluating,computer-use}.
Early computer-use methods include reactive agents \citep{yao2023reactsynergizingreasoningacting,xie2024osworldbenchmarkingmultimodalagents} that directly prompts an LLM (e.g., GPT-4o) to make decisions on immediate observations without simulation or planning approaches.
Recent advances include: 1) search-based methods \citep{zhou2024languageagenttreesearch,koh2024treesearchlanguagemodel,yu2024exactteachingaiagents} that augments LLMs with look-ahead search algorithms such as MCTS; and 2) hierarchical-planning methods that orchestrates multiple modules, tools, and LLMs to complete a task \citep{agashe2024agentsopenagentic,agashe2025agents2compositionalgeneralistspecialist,liu2025pcagenthierarchicalmultiagentcollaboration,gou2025navigatingdigitalworldhumans,yang2024ariauivisualgroundinggui}.
However, search-based methods significantly increase inference time due to using additional interactions with the external environment; and hierarchical-planning methods often require complex human-designed rules and heuristics to coordinate multiple modules.
We introduce our \internalDyna{} framework to augment the thinking process of a single LLM agent by performing action-centric world model simulation.

\paragraph{Training AI Agents}
Besides improving agent’s performance at test-time, many works also explored methods to improve performance via training.
Recent methods include \citet{chen2023fireactlanguageagentfinetuning,zhang2024xlamfamilylargeaction,zeng2023agenttuningenablinggeneralizedagent,lai2024autowebglmbootstrapreinforcelarge,xu2025aguvisunifiedpurevision} which perform supervised training using human or machine generated trajectories based on direct prompting; \citet{qin2025uitarspioneeringautomatedgui,su2025learnbyinteractdatacentricframeworkselfadaptive,hong2024cogagentvisuallanguagemodel} which improves agent's ability such as GUI grounding and error recovery via SFT/DPO on human-machine collaboration data; and \citet{bai2024digirltraininginthewilddevicecontrol,chen2025reinforcementlearninglonghorizoninteractive,jin2025searchr1trainingllmsreason}, which explores using direct RL to improve agent's policy within complex tool-use/android-based environments.
These methods focus on policy improvements based on an existing thinking paradigm (\eg ReACT or R1-style thinking).
We propose \wmreshort{} to improve the agent's reasoning ability by integrating world model simulation into its thinking process, and \internalDynatrainingshort{} to further enhance policy training via world model training.

\paragraph{World Models}
Obtaining real-world data for large-scale training or test-time search are expensive and may cause unintended consequences.
To this end, early methods such as \citet{peng2018deepdynaqintegratingplanning,wu2018switchbasedactivedeepdynaq,fang2025webevolverenhancingwebagent} consider Dyna-style training that separately trains a world model and then enhance policy training using synthetic rollouts; and prompting LLMs as world models \cite{hao2023reasoninglanguagemodelplanning,kim2024languagemodelsextrapolateoutside} in simplified environments such as BlocksWorld \citep{valmeekam2023planningabilitieslargelanguage}.
Recently, \citet{chae2025webagentsworldmodels,gu2025llmsecretlyworldmodel} trains a world model using a large corpus of web data to facilitate inference-time algorithms such as MCTS.
To our knowledge, this is the first work to propose internalizing world model simulation into the agent's thinking process, and to introduce methods to further improve the agent's policy by both world model and policy training.

\paragraph{Dyna Algorithms}
Dyna algorithms \citep{dyna} combine
 model-based and model-free methods to learn optimal policies. These methods improve training efficiency of \policy{} by combining real-world interaction with simulated planning.
Given a set of real-world rollout data, Dyna algorithms typically 1) separately trains a world model \worldmodelmu{} using these rollouts; 2) perform additional rollouts using \worldmodelmu{}; and 3) trains \policy{} using both the real-world and simulated rollouts.
Given a large set of training tasks to perform rollouts, this two-stage training process can be repeated for multiple times.
Applications of Dyna algorithms with language models include Deep Dyna-Q \citep{peng2018deepdynaqintegratingplanning}, Switch-DDQ \citep{wu2018switchbasedactivedeepdynaq}, Pseudo-DDQ \citep{pseudo-dyna-q} and more, covering domains such as movie-ticket booking and e-commerce product recommendations.

\section{\internalDyna{} Framework}
\label{sec:Method}

\input{floats/algo}

Accurate world simulation is challenging for AI agents.
We propose \internalDyna{}, a two-stage training method to address this.
First, we introduce \wmreshort{} (\Cref{subsec:Integrating World Modeling into Reasoning}) to synergize reasoning, acting, and world model simulation in an agent's thinking process via imitation learning. Then, we introduce \internalDynatrainingshort{} (\Cref{subsec:main_traing_method}) to further improve its policy and world modeling ability via Dyna-style training.

\subsection{Task Definition}
\label{subsec:Task Definition}
Completing tasks in a complex environment (\eg with a computer) is typically formulated as a partially observable Markov decision process (POMDP) of $(\mathcal{S}, \mathcal{O}, \mathcal{A}, \mathcal{T},\mathcal{R})$.
In the context of computer-use, $\mathcal{S}$ represents the set of possible computer states, $\mathcal{O}$ is the set of observations available to the agent, $\mathcal{A}$ represents the set of executable actions as left/right-clicks at a location, typing, and pressing keyboard shortcuts, $\mathcal{T}$ is the transition function $\mathcal{T}: \mathcal{S} \times \mathcal{A} \to \mathcal{S}$ that determines the next state given the current state and action, and $\mathcal{R}$ is the reward function that provides feedback to the agent based on its actions.
In typical computer-use benchmarks, $o \in \mathcal{O}$ is either a screenshot of the current computer screen or its text-based representation (\eg accessibility tree); and $r=0$ is zero if the task has not terminated, and $r_T=\{0,1\}$ if the agent failed/succeeded the task at the end.
Since $|\mathcal{A}|$ is extremely large and tasks such as computer-use often requires many steps to complete, many agent benchmarks remain challenging even for current state-of-the-art LLMs.

Given a task, a computer-use agent iteratively interact with the environment by generating an action $a_i$ based on the current observation $o_i$ and the past history of observations and actions $(o_{<i}, a_{<i})$.
In this work, we refer to such language model policies parametrized by $\theta$ as \policy{}, and learnt world models parametrized by $\mu$ as \worldmodelmu{}.
We denote policies that performs interaction with an internal world model during its thinking process as \roneminimal{}.

\subsection{\wmre{}}
\label{subsec:Integrating World Modeling into Reasoning}


Many expert LLMs capable of extensive thinking (e.g., DeepSeek R1, Claude-3.7-Sonnet, and OpenAI's o3) exhibits complex behaviors during its thinking process. This includes being able to perform verification, self-reflection, error recovery, world modeling, and more.
However, it is unclear which behaviors are critical for decision-making in long-horizon AI agent tasks.
In our preliminary study, we find training on long CoT data with simulations/knowledge unrelated to the final action degrades performance, a phenomenon also found in other domains \citep{zhou2024languagemodelsperformrobust,wu2025lessunderstandingchainofthoughtlength}.
To enable \internalDyna{} for ``weak'' non-reasoning models, we propose \textbf{\wmre{}} (\wmreshort) to first construct reasoning data that emphasizes on world simulation and perform imitation learning.
Specifically, \wmreshort{} reconstructs R1's thinking process to only contain text related to reasoning, the final action, and the \emph{world modeling simulation} related to the final action; and then trains a policy using the reconstructed data.
To perform this reconstruction, we few-shot prompt GPT-4o (see \Cref{sec:wm-distill-data} for examples and more details).
After training, we refer to these models as $\pi_{\mathcal{W}}(\theta)$, highlighting their ability to perform world-model simulations during thinking.


\subsection{\internalDynatraining{}}
\label{subsec:main_traing_method}

Despite \wmreshort{} learning, \roneminimal{} at inference time can still make mistakes when facing unseen states unseen from training.
However, creating a \wmreshort{} training set that covers all possible states $|\mathcal{S}|$ is intractable.
To further improve \wmreshort{}-trained models, we propose \textbf{\internalDynatraining{}} (\internalDynatrainingshort{}), a Dyna-style method that performs both policy and world model learning during online training.
Similar to Dyna, we first collect policy and world model learning data by performing \roneminimal{} rollouts in the real environment.
Different from Dyna, we then directly perform both policy and world model learning on the same \roneminimal{} model. We illustrate our method in \Cref{fig:all_algo}.

\paragraph{Policy Training} Alike reinforcement learning, the policy training stage aims to directly improve a policy using environment feedback.
This is achieved by first generating online rollouts with $\mathcal{W}$ to collect a trajectory $\tau = (o_0, a_1, o_1, a_2, ..., a_T)$, and then constructing a policy learning dataset that trains \roneminimal{} to predict each action $a_i$ given the previous context context $(o_{0}, a_{<i},o_{<i})$ based on a reward function (e.g., task success).
Optimizing such reward can be done using algorithms such as PPO \citep{schulman2017proximalpolicyoptimizationalgorithms}, GRPO \citep{shao2024deepseekmathpushinglimitsmathematical}, DPO \citep{rafailov2024directpreferenceoptimizationlanguage}, or Rejection Sampling \citep{bai2022constitutionalaiharmlessnessai,touvron2023llama2openfoundation}.

\input{floats/world_model_data}
\paragraph{World Model Training} Given a rollout trajectory $\tau$, we also construct world model training dataset to train \roneminimal{} to model a \emph{function} of the environment transition $\hat{\mathcal{T}_f}(o_i, a_i)$.
We experiment with three different functions for $\hat{\mathcal{T}_f}$ in this work.
$\hat{\mathcal{T}}(o_i,a_i) \to o_{i+1}$ which directly models the next state; $\hat{\mathcal{T}}_\Delta(o_i,a_i) \to \Delta (o_i,o_{i+1}|a_i)$ which trains to predict relevant \emph{changes} in the next state caused by $a_i$; and $\hat{\mathcal{T}}_{\textrm{critic}}(o_i,a_i) \to \textrm{critic}(o_i, a_{i} | o_{i+1})$ which trains to predict a \emph{critique} generated by an LLM (GPT-4o) by comparing the world model simulations in $a_i$ with the previous and next state.
An example world model simulation could be ``After opening the terminal with Ctrl+Alt+T, type `cp dir2/hi.txt dir1/' to copy ...'', and an example critique is ``Wait, maybe we need to first ensure that `dir1' exists in the current directory...''.
We illustrate these three functions in \Cref{fig:world_model_data}.
To utilize these objectives to \emph{enhance the world modeling ability of the policy model}, we optimize next-state prediction $\hat{\mathcal{T}}$ and state-changes prediction $\hat{\mathcal{T}}_\Delta$ as an ``auxiliary'' task trained by standard language modeling loss alongside policy training; and we optimize $\hat{\mathcal{T}}_{\textrm{critic}}$ by injecting the generated critique back into the action ($a_i' = a_i \oplus \textrm{critic}$) and then train the policy to predict critic tokens in $a_i'$ using language modeling loss.
Intuitively, $\hat{\mathcal{T}}$ and $\hat{\mathcal{T}}_\Delta$ enhance a policy's world modeling ability implicitly, while $\hat{\mathcal{T}}_{\textrm{critic}}$ is more explicit.
For more details, please see \Cref{subsec:world-model-data-prompts,subsec:internal-dyna-training-details}. 

Finally, we combine policy and world model training, \internalDynatrainingshort{} follows Dyna methods and perform  two-stage learning by first training \roneminimal{} on the world model dataset, and then training \roneminimal{} on the policy dataset.
For pseudocode, please see \Cref{alg:ddt}.
For other implementation details such as critique prompts and how training data is formatted, please refer to \Cref{sec:dyna-think-training-all-details}.

\section{Experiments}
\label{sec:Experiments}

\subsection{Benchmarks}
\label{subsec:Benchmarks}
We mainly evaluate our methods on OSWorld \citep{xie2024osworldbenchmarkingmultimodalagents}, a diverse benchmark consisting of 369 open-ended computer tasks that involves real web and desktop apps in open domains, OS file I/O, and workflow tasks. Tasks are categorized into 10 different domains based on different desktop applications involved, such as OS Terminal, LibreOffice Calc, LibreOffice Impress, LibreOffice Writer, Chrome, VLC Player,
Thunderbird, VS Code, GIMP and Workflow.

To evaluate the self-improving ability of AI agents, many methods \citep{huang2022largelanguagemodelsselfimprove,yu2024teachinglanguagemodelsselfimprove,chen2025reinforcementlearninglonghorizoninteractive,jin2025searchr1trainingllmsreason} consider domains where the model has non-trivial initial performance.
However, in OSWorld we find many domains such as LibreOffice Calc and Workflow to be extremely challenging, with current models achieving less than 10\% success rate or solving only 2-3 tasks often due to state representation issues (see \Cref{subsec:osworld-full-domain-perf}).
To this end, we evaluate our method on 5 domains that are more accessible to existing models, including OS, Chrome, VS Code, GIMP, and Thunderbird; and also \emph{additionally on WindowsAgentArena} \citep{bonatti2024windowsagentarenaevaluating} to further measure model's generalization ability to a different operating system (Windows OS).

\subsection{Experimental Setup}
\label{subsec:Experimental Setup}
\paragraph{Training/Testing Dataset}

Since most computer-use benchmarks were not designed for training, the number of tasks available in each domain is often limited.
We thus construct a training and test set by 1) manually augmenting existing tasks in each domain to increase its size; 2) construct training/test splits for each domain, and 3) held out two domains (GIMP and Thunderbird) from training to separately measure \textbf{In-domain (ID)} and \textbf{Out-of-domain (OOD)} performance.
To augment a task, we follow the principle that 1) action sequence that correctly completes task $A$ does \emph{not} complete the augmented task $A'$; and 2) the augmented task $A'$ can still be evaluated using OSWorld's evaluation scripts after some adjustments.
Please refer to \Cref{subsec:osworld-task-aug} for more details and example augmented tasks. In \Cref{tbl:train_test_set} we report the training and testing dataset statistics.
For WindowsAgentArena, we directly test on the official tasks as only OSWorld tasks were used in training.


\paragraph{Evaluation Details}
We evaluate all runs using the accessibility-tree mode (\ie text-only) on both benchmarks, and report task success for ID and OOD tasks.
We use the same inference prompts and hyperparameters (\eg temperature of 1.0) provided by OSWorld, but extend the maximum number of steps per task to 30 in order to measure scaling abilities.
To provide a more robust evaluation for these long-horizon tasks, we report the average success rate (Avg) and the Best-of-N success rate (BoN) over 3 runs.
Each evaluation is ran with 3 threads, lasting on average 18-24 hours to complete.

\paragraph{Training Details}
We train all of our models based on Qwen2.5-32B-Instruct \citep{qwen2025qwen25technicalreport} using 8xH100 GPUs.
We use 32B models due to the limited learning ability of smaller models for complex thinking patterns \citep{li2025smallmodelsstrugglelearn}.
For simplicity, we use rejection sampling as the optimization algorithm for policy training, and SFT for world model training.
For all runs, we use AdamW with a linear learning rate scheduler with 10\% warmup steps.
In the two-stage \internalDynatraining{}, we first perform world model training for 2 epoch with a max learning rate of 5e-6, and then continue training on policy learning data for 3 epochs with a max learning rate of 5e-6.

\subsection{Main Results}
\label{subsec:improve_wm_simluation_results}

\paragraph{Baselines}
We compare our \internalDyna{} framework against 1) training-free methods based on expert LLMs; and 2) related training methods.
For training-free methods, we consider prompting LLMs such as o3-mini and DeepSeek-R1 using \react{}.
For training-based methods, we consider 1) \textbf{\rftlong{}} (\rft{}) which only peforms policy learning, by finetuning on correct rollouts after rejection sampling (Best-of-3);
and 
2) the vanilla \textbf{Dyna} algorithm (we follow \citet{fang2025webevolverenhancingwebagent}) which performs world model learning with a separate language model \worldmodelmu{}, and then trains the policy using correct rollouts\footnote{When rolling out with the learned \worldmodelmu{}, correctness of a trajectory is evaluated using an LLM \citep{fang2025webevolverenhancingwebagent}. We use GPT-4o in this work.} collected with \emph{both} the real $\mathcal{W}$ and the learned \worldmodelmu{}.
Our \internalDynatrainingshort{} method only uses rollouts with $\mathcal{W}$ and performs policy and world model learning on a single model.

For a fair comparison, we use the same set of rollout trajectories (116 in total) to construct world model (116) and policy training data (35 after rejection sampling, a similar data size as \cite{yu2024exactteachingaiagents}) for all training methods, whenever possible.

\input{floats/world_model_v2}
\input{floats/winarena_success}
\paragraph{Results} We report our results on OSWorld in \Cref{tbl:wm_v2}, and results on WindowsAgentArena on \Cref{tbl:winarena_results}.
We first compare training-based methods.
In \Cref{tbl:wm_v2}, we find that \internalDynatrainingshort{} training, especially when trained with next-state prediction (\wmprefinetuneshort{}) and critique prediction (\judgeprefinetuneshort{}) improves upon both \rft{} and vanilla Dyna in both average success rate and BoN success rate.
This indicates that 1) world model training benefits policy training; and 2) directly training \roneminimal{} as a world model is more effective than planning with a separately trained \worldmodelmu{}.
Next, we find that training with critique data showed strong performance compared to training on state-difference (\wmdiffprefinetuneshort{}) and on next-state prediction (\wmprefinetuneshort{}).
We believe this is because these critique data provide a more direct signal for \roneminimal{} to improve its world modeling and planning ability \emph{during inference} (see \Cref{subsec:World Model Accuracy Test} for more quantitative study).

Moreover, we find that training with the next-state prediction objective (\wmprefinetuneshort{}) shows strong performance for OOD tasks, especially on WindowsAgentArena.
We believe this is because next-state prediction more effectively enhanced the model’s understanding of the deskptop environments, which is useful for OOD where many states are entirely novel.
Finally, compared to best training-free methods such as prompting R1(685B), we find \internalDyna{} achieves similar BoN score, while generating 2x less tokens on average and being only 32B in size.
These results indicate the many tokens/behaviors during R1-style reasoning may not be necessary, and that \emph{focusing on/improving simulation ability} is effective at improving agent performance.

\input{floats/enable_v1}
\subsection{Thinking Behavior Analysis}
\label{subsec:integrate_wm_in_reasoning_results}

We now investigate what behavior is essential for long-horizon AI agent tasks.
We compares agents with no-thinking; R1-style thinking, and \internalDyna{}.
For \textbf{no-thinking}, we consider 1) \directdistillshort{}(4o) which distills from GPT-4o that, to our observation, does \emph{not} perform world modeling \citep{chae2025webagentsworldmodels};
and 2) \directdistillshort{}(R1 no-think) which trains only on tokens after the thinking process (by removing all text within the `<think></think>' tags in each response).
For \textbf{R1-style} thinking, we consider \directdistillshort{}(R1), which trains on the entire thinking process generated by R1.
For \textbf{\internalDyna{}}, we consider \wmreshort{}(R1).
For a fair comparison across different methods, we use the \emph{same set of correct trajectories} obtained by best-of-3 rejection sampling.

We present the results in \Cref{tbl:enable_v1}.
First, we find that \directdistillshort{}(R1 no-think) deteriorates significantly compared to \directdistillshort{}(R1), even though they are trained on the same set of trajectories.
This indicates that including long-CoT thinking during policy training is benefitial.
Next, we find that \wmreshort{}(R1) achieved similar performance compared to \directdistillshort{}(R1), despite generating 2x less tokens on average.
This suggests that the ability to perform \emph{reasoning with world-model simulations} is the central part in R1-style thinking, underscoring the importance of world modeling in AI agents.

\vspace{-2mm}
\section{Discussions}
\label{sec:Discussion and Challenges}

In this section, we study if the ``upperbound'' of our model's performance  (\eg Best-of-3 success rate) can be further improved via 1) scaling world model training data; and 2) iteratively bootstrapping ``better'' policy data.
Finally, we quantatively measure different model's world model ability, and compare it with their overall performance.
\subsection{Scaling World Model Training in \internalDyna{}}
\label{subsec:scaling_wm_judge_data}

\input{floats/scaling_v1fig}

Since world model learning only requires interaction data (without evaluators for task success) with the real $\mathcal{W}$, we investigate whether scaling world training in \Cref{subsec:main_traing_method} can further improve policy performance.
To test this, we 1) prompt GPT-4o to generate synthetic task instructions for each domain; 2) use \roneminimal{} collect rollout trajectories; and 3) construct a world model learning dataset following \Cref{subsec:main_traing_method}.
Since no evaluator is required for world modeling data construction, we used all rollout trajectories that terminated within a maximum of 30 steps.
This results in a total of 703 additional trajectories for world model learning.
For more details please see \Cref{subsec:sythnetic-task-gen-for-wm}.

We present our results in \Cref{fig:scaling_wm_fig}.
In \Cref{fig:scaling_wm_results}, we find that training with additional world model data steadily improves model's BoN success rate.
This indicates that scaling world model training (with synthetic tasks) enhances the agent’s environment understanding, enabling it to solve novel tasks.
However, we did not observe a substantial increase its ``robustness'' - the agent did not consistently solve these novel tasks across all three trials.
We believe this may be due to the stochastic nature of real-world environments, and that increasing world model training data alone cannot ensure the agent to robustly utilize all relevant world knowledge in its policy.
We suggest future work should scale both world model and policy training data together, by 1) manually creating a large set of agent tasks with evaluators available for training; and/or 2) developing robust automatic evaluators \citep{pan2024autonomousevaluationrefinementdigital} capable of evaluating synthetic tasks generated on-the-fly.



\subsection{Iterating Policy Training in \internalDyna{}}
\label{subsec:Iterative Training Results}

We now investigate whether \roneminimal{} can be iteratively trained \emph{without} any supervision from an expert LLM (\eg GPT-4o was used to critique simulation in \Cref{subsec:main_traing_method}).
Specifically, we follow STaR (\citet{zelikman2022starbootstrappingreasoningreasoning}, a simple method use for math and reasoning domain) and consider two iterative training loops.
\textbf{Without Evaluation Hint} (w/o eval hint), where we 1) first perform rejection sampling using \roneminimal{} on the training set; 2) perform policy learning on \roneminimal{}; and 3) repeat.
\textbf{With Evaluation Hint} (w/ eval hint), where for tasks that \roneminimal{} fail to solve during step 1, we perform ``rationalization'' by appending the evaluation configuration to the original instruction (\Cref{fig:rationalization_expl}), and then perform rejection sampling again on these tasks with evaluation hint.
During training and testing, we remove the added hints from the instruction.
We report test performance over 5 training iterations.

\input{floats/iterative_training_v1fig}




We present our results in \Cref{fig:iterative_policy_results}.
We find iterative training without rationalization (w/o eval hint) quickly plateaus; and that training with rationalization (w/ eval hint) outperforms training without rationalization.
However, even when provided with evaluation configurations, we observe that \roneminimal{} can only solve a portion of the tasks that it fails to solve otherwise.
This indicates that agent tasks, such as computer-use, remains to be a challenging domain for current language models.

\subsection{Quantifying World Model Accuracy}
\label{subsec:World Model Accuracy Test}

\input{floats/wm_acc_v1}
In this work, we introduced \internalDyna{} to integrate world model learning with policy learning in a single \roneminimal{}.
To measure the effectiveness of this world model learning, we now evaluate the \textbf{World Model Accuracy} (Acc.) of different models, and the \textbf{Pearson Correlation Coefficient} ($r$) between the (average) world model accuracy for each task and the task success.
To evaluate world model accuracy given an action $a_i=\langle \textrm{think}_i, \textrm{act}_i \rangle$ generated by \roneminimal{}, we 1) prompt GPT-4o to extract the world model simulation text from $\textrm{think}_i$ that corresponds to $\textrm{act}_i$; and 2) prompt GPT-4o judge whether the extracted simulation is correct \emph{given the next state $o_{i+1}$} after executing $a_i$.
For each model, we calculate this for every turn in each trajectory that terminated within a maximum of 30 steps.
Please refer to \Cref{subsec:wm-acc-test-details} for prompts used and more details.

In \Cref{tbl:wm_acc_v1.1}, we find that 1) models that achieve a higher success rate also achieve a higher world model accuracy, and that 2) average world model accuracy for each task shows strong correlation with task success, with a minimum correlation of $r=0.32$ across all models.
Next, we find that \internalDynatrainingshort{} training significantly improved the world model accuracy (16.8\% absolute) along with an improved success rate, even though it was trained on the \emph{same set of policy data} as RFT (see \Cref{subsec:improve_wm_simluation_results}).
This shows that combining world-model and policy learning effectively boosts AI agent performance.






\vspace{-1mm}
\section{Conclusion}
\label{sec:Conclusion}
\vspace{-1mm}

We present \internalDyna{}, a new thinking framework that synergizes reasoning, acting, and \emph{planning by simulating with an internal world model} to improve the performance of AI agents.
\internalDyna{} consists of two training stages: \wmreshort{} to initialize a model with simulation ability during reasoning, and \internalDynatrainingshort{} for further improvement.
We evaluated our methods on OSWorld and WindowsAgentArena, and find our models based on Qwen2.5-32B-Instruct reach a similar best-of-n performance compared to R1(685B), while generating 2x less tokens on average.
Our empirical analysis reveals that 1) critique-style world model training is effective for policy improvement; and 2) stronger AI agents show stronger world modeling ability. Our results suggest a promising direction for integrating world model simulation into AI agents to enhance their reasoning, planning, and acting abilities.

\bibliography{iclr2026_conference}
\bibliographystyle{iclr2026_conference}

\clearpage
\appendix

\setcounter{table}{0}
\renewcommand{\thetable}{A\arabic{table}}
\setcounter{figure}{0}
\renewcommand{\thefigure}{A\arabic{figure}}

\section{LLM Usage}
This work used LLMs as general-purpose writing assistants to improve the grammar and clarity of the paper.
We DO NOT use LLMs to generate research ideas, automate experiments, or analyze results.

\section{Limitations}
\label{sec:Limitations}

\paragraph{Training Long-Horizon Trajectories}
Computer-use tasks on benchmarks such as OSWorld \citep{xie2024osworldbenchmarkingmultimodalagents} often require 10 and sometimes up to 100 steps to complete using current LLM \citep{openai-cua,computer-use}.
This makes model training challenging due to significantly increased sequence length.
In this work, we train each $a_i$ by keeping a maximum context length of 20480 tokens (about 3-4 turns), and use DeepSpeed Zero-3 \citep{rajbhandari2020zeromemoryoptimizationstraining} to reduce memory usage. In general, we believe memory-efficient methods to enable long-context training will be beneficial, which we leave for future work.


\paragraph{Base Model Capability}
Powerful agents based on LLMs such as DeepSeek-R1 often have model sizes reaching hundreds of billions of parameters.
In this work, we used Qwen2.5-32B-Instruct as we found training smaller models (\eg 7B in size) yields limited performance improvements.
We believe this is due to the challenging nature of 1) computer-use tasks; and 2) learning long-CoT data \citep{li2025smallmodelsstrugglelearn}.
To significantly improve computer-use performance for these smaller models, we believe substantial task-specific post-training may be required. We leave this for future work.


\vspace{-2mm}
\section{Ethics Statement}
\label{label:ethics statement}
Generally, while most methods and models are not designed for unethical usage, there is often potential for abuse in their applications.
Computer-use agents can be used for a wide-range of tasks such as automating form filling, information gathering, software development, and more.
In this work, we proposed our \internalDyna{} framework to enhance the performance and (token) efficiency of AI agents.
However, since computer-use agents are fundamentally task-agnostic, it is possible to use them for unethical tasks such as scamming or disseminating false information on the internet.
We believe developing guardrails such as safety filters \citep{openai-content-filters,inan2023llamaguardllmbasedinputoutput} are highly valuable for AI agent research.
We do not condone the use our \internalDyna{} methods for any unlawful or morally unjust purposes.

\vspace{-2mm}
\section{\wmre{} Details}
\label{sec:wm-distill-data}

\input{floats/minimal_text_example}
We use GPT-4o to reconstruct R1-style reasoning in \Cref{subsec:Integrating World Modeling into Reasoning}.
For each turn in an R1-generated trajectory, we prompt GPT-4o to reconstruct $a_i$ given its immediate previous observation $o_{i-1}$ and relevant system instructions.
We present our prompt in \Cref{tbl:wm_distill_prompt}.
A reconstructed output is shown in \Cref{fig:minimal_example}.
For R1 responses within 550 tokens, we do not perform abbreviation as these responses already mostly only constitutes verification, world model simulation, and action.
To help ensure GPT-4o removes as much token as possible, for each prompt we sample 5 outputs, and use the shortest output in our \wmreshort{} training corpus.

\section{\internalDynatraining{} Details}
\label{sec:dyna-think-training-all-details}

\subsection{World Model Data Prompts}
\label{subsec:world-model-data-prompts}

To construct world model data, we experimented with three methods in \Cref{subsec:main_traing_method}.
\textbf{Next-state prediction} (\wmprefinetuneshort{}) which trains the model to directly predict the next state; \textbf{State-difference prediction} (\wmdiffprefinetuneshort{}) which trains the model to predict the difference $\Delta(o_{i},o_{i+1})$; and \textbf{Simulation-critique generation} (\judgeprefinetuneshort{}), which trains the model to generate a critique for simulation in $a_i$.
To obtain data for \wmdiffprefinetuneshort{} we prompt GPT-4o using prompts in \Cref{tbl:wm_diff_prompt}.
To obtain data for \judgeprefinetuneshort{}, we first prompt GPT-4o to extract the world model simulation in $a_i$ corresponding to its final action, and the prompt it to generate a critique using prompts in \Cref{tbl:wm_judge_prompt}.
Example next-state, state-difference, and simulation-critique data is shown in \Cref{fig:wm_data_example}.

\subsection{Training Details}
\label{subsec:internal-dyna-training-details}
\input{floats/wm_judge_example}
We present an overview of \internalDynatraining{} in \Cref{fig:full_training_pipeline}.
Given a set of rollout trajectories, we first perform world model training and then perform policy training.
During policy training, the model \roneminimal{} is trained to predict the next action $a_i$ given its previous context using correct trajectories.
During world model training, we experiment with three data formats:
\wmprefinetuneshort{} directly trains the model to predict the next state $o_{i+1}$; \wmdiffprefinetuneshort{} trains the model to predict a natural language description of the state difference $\Delta({o_i, o_{i+1}})$ generated by prompting GPT-4o (see \Cref{tbl:wm_diff_prompt}; and \judgeprefinetuneshort{} trains the model to generate a critique for the world model simulation in $a_i$ that caused $o_{i+1}$ (see \Cref{tbl:wm_judge_prompt}).

To better localize the critique to its corresponding world model simulation, we additionally 1) prompt GPT-4o to inject the critique back to the original response; and 2) only train on the critique by masking out all other tokens (see \Cref{fig:judge_data_example}).
To inject a critique back into $a_i$, we 1) append an id ``[[id=x]]'' at the end of every sentence in $a_i$; and 2) prompt GPT-4o to output an injection location; and 3) inject the critique back into the response based on the output id.

In \Cref{alg:ddt} we provide a high-level pseudocode for \internalDynatrainingshort{}. For world model training, we adopt Language Modeling(LM) Loss:
\begin{equation}
\mathcal{L}_{\text{LM}}(\theta) 
= - \sum_{t=1}^T \log \pi(x_t \mid x_{<t}),
\label{eq:lm-loss}
\end{equation}

During policy training, we use policy gradient to optimize the reward: 
\begin{equation}
\nabla_\theta\mathcal{L}_{\text{policy}}(\theta) 
= \mathbb{E}_{\tau\sim\pi_{\mathcal{W}}(\theta)}
\Big[ \sum_{t=0}^T\nabla_\theta\log \pi(a_t\mid o_t) R(\tau) \Big],
\label{reinforce-formula}
\end{equation}
where we use $R(\tau)=1$ for successful tasks, otherwise $0$.
In our main experiment (\Cref{tbl:wm_v2}), we perform $N=1$ iteration enable to have a fair comparison between the data used for policy and world model training across different algorithms (RFT and vanilla Dyna).
In \Cref{subsec:Iterative Training Results} we extend our method to multiple iterations.

\algdef{SE}[REPEATN]{RepeatN}{End}[1]{\algorithmicrepeat\ #1 \textbf{times}}{\algorithmicend}
\begin{algorithm}[t]
\caption{DDT Training (Policy + World Model with Critique Injection)}
\label{alg:ddt}
\begin{algorithmic}[1]
\Require Policy/world model $\pi_W(\theta)$, environment $\mathcal{T}$, rollout trajectory $\tau$, reward function $R$
\RepeatN{$N$}
    \Comment{Data Collection:}
    \State Roll out $\pi_{\mathcal{W}}(\theta)$ in $\mathcal{T}$ for trajectory $\tau= (o_0, a_1, o_1, a_2, \ldots, a_T)$
    \State Construct policy dataset $\mathcal{D}_\pi \gets \{(\text{context}(o_0,a_{<i},o_{<i}), a_i, R)\}$
    \State Construct world model dataset $\mathcal{D}_W \gets \{(o_i, a_i, o_{i+1})\}$

    \Comment{World Model Training:}
    \ForAll{$(o_i, a_i, o_{i+1}) \in \mathcal{D}_W$}
    \State \textbf{Option 1:} Next-state prediction (DDT($\hat{\mathcal{T}}$))  
    \Statex \hspace{2.8em} Input $(o_i, a_i)$, target $o_{i+1}$  
    \Statex \hspace{2.8em} $\mathcal{L}_{\text{wm}} \gets \mathcal{L}_{\text{LM}}(\pi_W(\theta), (o_i,a_i), o_{i+1})$

    \State \textbf{Option 2:} State-change prediction (DDT($\hat{\mathcal{T}}_{\Delta}$))  
    \Statex \hspace{2.8em} Input $(o_i, a_i)$, target $\Delta(o_i,o_{i+1})$ (NL description from GPT-4o)  
    \Statex \hspace{2.8em} $\mathcal{L}_{\text{wm}} \gets \mathcal{L}_{\text{LM}}(\pi_W(\theta), (o_i,a_i), \Delta(o_i,o_{i+1}))$

    \State \textbf{Option 3:} Critique prediction (DDT($\hat{\mathcal{T}}_{\text{critic}}$))  
    \Statex \hspace{2.8em} (1) Append IDs \texttt{[[id=x]]} to each sentence in $a_i$  
    \Statex \hspace{2.8em} (2) Use GPT-4o to output injection location  
    \Statex \hspace{2.8em} (3) Inject critique $\to a'_i$  
    \Statex \hspace{2.8em} (4) Mask non-critique tokens in $a'_i$  
    \Statex \hspace{2.8em} $\mathcal{L}_{\text{wm}} \gets \mathcal{L}_{\text{LM}}(\pi_W(\theta), (o_i,\ldots), \text{masked critique tokens})$
\EndFor
    \State Update $\theta \gets \theta - \eta \nabla_\theta L_{\text{wm}}$

    \Comment{Policy Training:}
    \ForAll{$(\text{context}, a_i, R) \in \mathcal{D}_\pi$}
        \State $\mathcal{L}_{\pi} \gets \mathcal{L}_{\text{policy}}(\pi_W(\theta), \text{context}, a_i; R)$
    \EndFor
    \State Update $\theta \gets \theta - \eta \nabla_\theta \mathcal{L}_{\pi}$
\End

\State \Return $\pi_{\mathcal{W}}(\theta)$
\end{algorithmic}
\end{algorithm}

\subsection{Synthetic Task Generation}
\label{subsec:sythnetic-task-gen-for-wm}

Since world model training only requires rollout trajectories (regardless of their correctness), we experiment with scaling up world model training by using GPT-4o to generate synthetic tasks for rollouts.
Specifically, we prompted GPT-4o using the template in \Cref{tbl:task_generation_prompt}, and generated 200 tasks for each domain.
To ensure diversity between tasks, we prompt GPT-4o to directly generate $N=10$ distinct tasks in a single prompt, and repeat this process 20 times for each domain.
After this generation process, we manually inspected multiple task and did not find obvious duplicates or unreasonable tasks.

In practice, we notice that many tasks require additional configuration (\eg setting up email profile for the Thunderbird domain, or downloading an image to edit in GIMP).
To accommodate this, we consider a few domain-specific changes.
For all synthetic tasks in the thunderbird domain, we use the same email account configurations as used in the original OSWorld dataset.
For the GIMP domain, we augmented our prompt template in \Cref{tbl:task_generation_prompt} to additionally include an example GIMP task configuration from the OSWorld dataset, and instructed the model to generate new tasks based on the given configuration (see \Cref{tbl:gimp_generation_prompt}).
We then use the same task configuration used in the prompt, but replace the instruction with our generated task.
This process ensures that the agent has the necessary initialization for the task (\eg at least an image to edit).
For all other domains (OS, Chrome, VSCode), we use an empty configuration as they do not require additional resources.

Given these synthetic tasks, we directly perform rollouts using \roneminimal{}. We then remove trajectories that did not terminate within a maximum of 30 steps, and use all the remaining rollouts to construct world model training data used in \Cref{subsec:scaling_wm_judge_data}.
This resulted in a total of 6467 turns available for training, 8x more than the world model training data used in \Cref{subsec:improve_wm_simluation_results}.

\section{OSWorld Details}
\label{sec:osworld-details}

\subsection{Additional Domains}
\label{subsec:osworld-full-domain-perf}

In \Cref{tbl:full_domain_results}, we present our model performance on domains not included in our main except (VLC, LibreOffice Impress, LibreOffice Writer, LibreOffice Calc, and Workflow).
While we did observe improvement on average after training, we find these are limited (\eg 6\% improvement in VLC only corresponds to \emph{completing 1 additional task}).
Overall, we find solving tasks in these domains often encounters representation issue, such as the need to interact with a video (in VLC) or making \emph{visual} changes to a powerpoint slide or document (in LibreOffice tasks).
These visual information were often missing or mis-represented in the text-modality (accessibility tree) provided by the OSWorld benchmark.
Thus, we mainly compare methods on domains such as OS, VSCode, Chrome, GIMP, and Thunderbird, where most tasks are solvable under the text-modality.

\subsection{Task Augmentations}
\label{subsec:osworld-task-aug}

\input{floats/wm_training_output_example}

\input{floats/train_test_set}
As of the date of this work, OSWorld \citep{xie2024osworldbenchmarkingmultimodalagents} is the largest-scale computer-use benchmark available with 369 tasks covering 10 domains.
However, as creating evaluators for computer-use tasks is non-trivial, this benchmark mainly serves as testing performances due to its limited size for training.
To this end, we follow recent work \citep{su2025learnbyinteractdatacentricframeworkselfadaptive}  and increase the number of computer-use tasks.
Specifically, we manually create new tasks by modifying the original instruction; and/or the task initialization configuration; and the corresponding evaluator configuration.
We ensure the resulting task is 1) \emph{distinct} from the original task, such that action sequences that can correctly complete the original task does not complete the augmented tasks; and 2) \emph{can be correctly evaluated} by modifying the existing evaluator configurations.
We present some example augmented tasks in \Cref{tbl:test_aug_task_examples}.

\section{Analysis Details}
\label{sec:analysis-details}

\subsection{World Model Accuracy Test}
\label{subsec:wm-acc-test-details}

Given a trajectory from \roneminimal{} that performs world model simulation during its thinking process, we 1) first prompt GPT-4o to extract the world model simulation corresponding to the final action $a_i$; and then 2) prompt GPT-4o to judge the correctness of this simulation \emph{given the next state}. We repeat this for every $a_i$ in a all trajectories that terminated within a maximum budget of 30 steps.
We present the prompts used to extract world model simulation in \Cref{tbl:wm_acc_simextract_prompt}, and prompts used to measure simulation accuracy in \Cref{tbl:wm_acc_test_prompt}.

\input{floats/full_training_pipeline}
\input{floats/full_domain_results}
\input{floats/test_aug_tasks}

\input{floats/wm_distill_prompt}
\input{floats/diff_n_judge_prompt}
\input{floats/task_gen_prompt}
\input{floats/wm_acc_prompt}

\end{document}

%% file: math_commands.tex
\usepackage{amsmath,amsfonts,bm}

\def\eqref#1{equation~\ref{#1}}

\def\1{\bm{1}}

\DeclareMathAlphabet{\mathsfit}{\encodingdefault}{\sfdefault}{m}{sl}
\SetMathAlphabet{\mathsfit}{bold}{\encodingdefault}{\sfdefault}{bx}{n}



%% file: floats/dyna_think_example.tex
\begin{figure}[t!]
    \centering
    \includegraphics[scale=0.67]{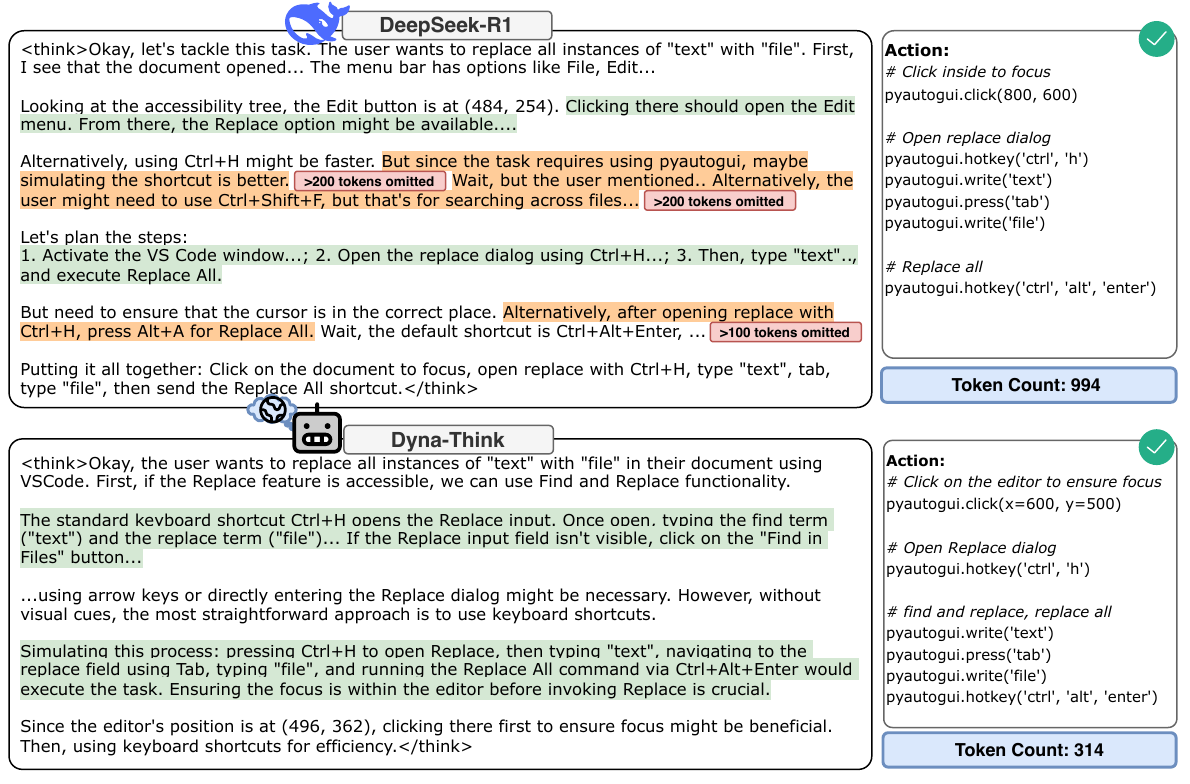}
    \caption{\internalDyna{} focus on integrating world model simulation (shown in \textcolor{teal}{green}) with reasoning and acting.
    Behaviors that is not necessary to cause/has unclear contribution to the final action is shown in \textcolor{orange}{orange}.
    After training, we find \internalDyna{} achieves similar BoN performance compared to R1, while generating 2x less tokens on average and being only 32B in size.
    }
    \label{fig:dyna_think_example}
\end{figure}

%% file: floats/algo.tex
\begin{figure}[t!]
    \begin{subfigure}[b]{0.328\textwidth}
        \centering
        \includegraphics[width=\linewidth]{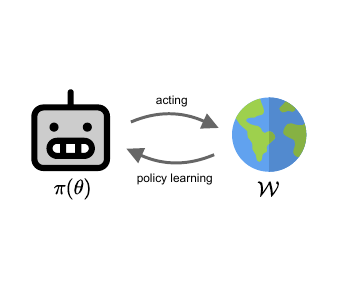}
        \caption{Reinforcement Learning}
        \label{fig:cmp_rft}
    \end{subfigure}
    \hfill
    \begin{subfigure}[b]{0.328\textwidth}
        \centering
        \includegraphics[width=\linewidth]{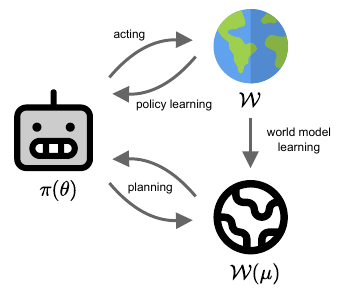}
        \caption{Dyna Training}
        \label{fig:cmp_dyna}
    \end{subfigure}
    \hfill
    \begin{subfigure}[b]{0.328\textwidth}
        \centering
        \includegraphics[width=\linewidth]{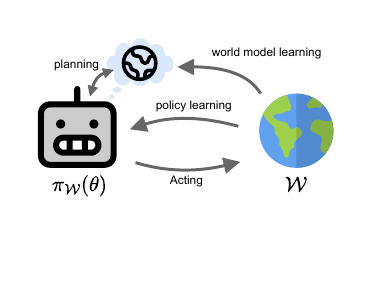}
        \caption{\internalDyna{} Training}
        \label{fig:our_algo}
    \end{subfigure}
\caption{
Our \internalDyna{} framework synergizes planning with world model simulation in an agent's reasoning process, and performs both world model training and policy training with \roneminimal{}.
}
\label{fig:all_algo}
\end{figure}

%% file: floats/world_model_data.tex
\begin{wrapfigure}[10]{R}{0.44\textwidth}
\vspace{-13mm}
    \centering
    \includegraphics[scale=0.64]{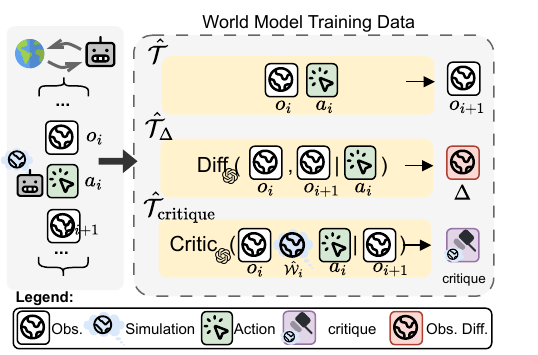}
    \caption{Three different forms of world model data experimented in \internalDynatrainingshort{}.}
    \label{fig:world_model_data}
\end{wrapfigure}

%% file: floats/world_model_v2.tex
\begin{table}[t!]
  \centering
  \caption{Average (AVG) and Best-of-N (BoN) success rate on OSWorld after policy and world model learning. $|\pi|$ and $|\mathcal{W}|$ denotes the number of trajectories used during policy learning and world model learning, respectively. ``Gen. Token'' denotes the 10th and 90th percentiles of output token lengths per prompt generated by the models. All training are based on Qwen-2.5-32B-Instruct. All methods are evaluated over 3 runs.
  For overall performance (All), we report average success $\pm$ standard deviation.
  }
  \vspace{-3mm}
  \scalebox{0.76}{
  \begin{tabular}{l l l l ccc ccc}
    \toprule
    \multirow{2}{*}{Method} & 
    \multirow{2}{*}{$|\pi|$} & 
    \multirow{2}{*}{$|\mathcal{W}|$} & 
    \multirow{2}{*}{Gen. Token} & 
    \multicolumn{3}{c}{Avg Success Rate} &
    \multicolumn{3}{c}{BoN Success Rate}
    \\
    \cmidrule(lr){5-7}
    \cmidrule(lr){8-10}
    & & & 
    (10\%-90\%) &
    \multicolumn{1}{c}{All(174)} &
    \multicolumn{1}{c}{ID(123)} &
    \multicolumn{1}{c}{OOD(51)} &
    \multicolumn{1}{c}{All(174)} &
    \multicolumn{1}{c}{ID(123)} &
    \multicolumn{1}{c}{OOD(51)} \\
    \midrule

    \rowcolor{light-gray}
    - (GPT-4o-2024-11-20) & - & -
    & 1.0x
    & 16.5\small{$\pm$2.5} & 19.5 & 9.2
    & 31.6 & 36.6 & 19.6 \\

    \rowcolor{light-gray}
    - (Qwen2.5-32B-Instruct) & - & -
    & 0.2x-0.8x
    & 14.2\small{$\pm$2.2} & 18.9 & 2.5
    & 27.0 & 35.0 & 7.8 \\

    \rowcolor{light-gray}
    - (R1-distill-70B) & - & -
    & 0.6x-2.2x
    & 7.6\small{$\pm$2.0} & 9.8 & 2.5
    & 15.5 & 20.3 & 3.9 \\

    \rowcolor{light-gray}
    - (o3-mini-2025-01-31) & - & -
    & 3.4x-7.0x
    & 20.5\small{$\pm$1.4} & 23.3 & 13.7
    & 31.0 & 36.6 & 17.6 \\

    \rowcolor{light-gray}
    - (R1) & - & -
    & 1.7x-4.9x
    & 31.2\small{$\pm$1.4} & 35.8 & 20.2
    & 44.8 & 48.8 & 35.3 \\
    \midrule
    \wmreshort{}(R1) & - & -
    & 1.1x-2.5x
    & 22.6\small{$\pm$1.0} & 26.8 & 12.4
    & 35.6 & 40.7 & 23.5 \\
    
    \quad +\rft{} & 35 & -
    & 1.0x-2.7x
    & 23.2\small{$\pm$1.0} & 26.0 & 16.3
    & 38.5 & 43.1 & 27.5 \\

    \quad +vanilla Dyna & 94 & 116
    & 1.1x-2.5x
    & 24.1\small{$\pm$2.1} & 26.8 & 17.6
    & 35.6 & 29.3 & 31.4 \\
    
    \midrule
    \quad +\wmprefinetuneshort{} & 35 & 116
    & 1.1x-2.7x
    & 25.9\small{$\pm$1.4} & 28.7 & \textbf{19.0}
    & 43.1 & 47.2 & \textbf{33.3} \\
    
    \quad +\wmdiffprefinetuneshort{} & 35 & 116
    & 1.2x-3.5x
    & 23.2\small{$\pm$0.7} & 27.6 & 12.4
    & 38.5 & 44.7 & 23.5 \\
    
    \quad +\judgeprefinetuneshort{} & 35 & 116
    & 1.2x-2.7x
    & \textbf{26.6}\small{$\pm$1.5} & \textbf{30.3} & 17.6
    & \textbf{44.3} & \textbf{49.6} & 31.4 \\
    \bottomrule
\end{tabular}
  }
  \label{tbl:wm_v2}
\end{table}

%% file: floats/winarena_success.tex
  

\begin{wraptable}[11]{R}{0.38\textwidth}
    \centering
    \caption{WindowsAgentArena results.}
    \label{tbl:winarena_results}
\scalebox{0.86}{
    \begin{minipage}{0.65\textwidth}
        \begin{tabular}{l c c}
            \toprule
            Method & Avg Success Rate \\
            \midrule
            \rowcolor{light-gray}
            - (Qwen-32B) & 23.9\small{$\pm$1.1} \\
            \rowcolor{light-gray}
            - (R1) & 26.9\small{$\pm$1.5} \\
            \midrule
            DIT(R1) & 26.9\small{$\pm$1.3} \\
            \quad + RFT & 28.4\small{$\pm$1.8} \\
            \quad + vanilla Dyna & 20.9\small{$\pm$1.1} \\
            \midrule
            \quad +\wmprefinetuneshort{} & \textbf{34.9}\small{$\pm$1.4} \\
            \quad +\judgeprefinetuneshort{} & 32.8\small{$\pm$1.4} \\
            \bottomrule
        \end{tabular}
    \end{minipage}
}
\end{wraptable}

%% file: floats/enable_v1.tex
\begin{table}[t!]
  \centering
  \caption{Integrating world model simulation (WM Sim) into reasoning.
  All training are based on Qwen2.5-32B-Instruct. ``R1 no-think'' refers to only training on tokens \emph{after} the ``</think>'' tag.
  Since o3-mini API does not return the model's thinking process, it is unclear if it performs world modeling.
  }
  \vspace{-3mm}
  \scalebox{0.72}{
  \begin{tabular}{l c l ccc ccc}
    \toprule
    \multirow{2}{*}{Method} & 
    \multirow{2}{*}{WM Sim?} & 
    \multirow{2}{*}{Gen. Token} & 
    \multicolumn{3}{c}{Avg Success Rate} &
    \multicolumn{3}{c}{BoN Success Rate}
    \\
    \cmidrule(lr){4-6}
    \cmidrule(lr){7-9}
    & & (10\%-90\%) &
    All(174) & ID(123) & OOD(51) &
    All(174) & ID(123) & OOD(51) \\
    \midrule

    \rowcolor{light-gray}
    - (GPT-4o-2024-11-20) & \xmark & 1.0x
    & 16.5\small{$\pm$2.5} & 19.5 & 9.2
    & 31.6 & 36.6 & 19.6 \\

    \rowcolor{light-gray}
    - (Qwen2.5-32B-Instruct) & \xmark & 0.2x-0.8x
    & 14.2{$\pm$2.2} & 18.9 & 2.5
    & 27.0 & 35.0 & 7.8 \\
    
    \rowcolor{light-gray}
    - (R1) & \cmark & 1.7x-4.9x
    & 31.2{$\pm$1.4} & 35.8 & 20.2
    & 44.8 & 48.8 & 35.3 \\
    
    \midrule
    
    \directdistillshort{}(4o) & \xmark & 0.5x-1.7x
    & 15.3\small{$\pm$2.4} & 18.5 & 7.8
    & 28.7 & 34.1 & 15.7 \\
    
    \directdistillshort{}(R1 no-think) & \xmark & 0.1x-0.5x
    & 17.1\small{$\pm$2.1} & 19.3 & 11.8
    & 28.2 & 30.9 & 21.6 \\
    
    \directdistillshort{}(R1) & \cmark & 1.6x-6.0x
    & 20.9\small{$\pm$1.0} & 24.6 & 11.8
    & \textbf{36.2} & \textbf{42.3} & 21.6 \\
    
    \wmreshort{}(R1) & \cmark & 1.1x-2.5x
    & \textbf{22.6}{$\pm$1.0} & \textbf{26.8} & \textbf{12.4}
    & 35.6 & 40.7 & \textbf{23.5} \\
    
    \bottomrule
\end{tabular}
  }
\vspace{-4mm}
\label{tbl:enable_v1}
\end{table}

%% file: floats/scaling_v1fig.tex
\begin{figure}[t!]
    \begin{subfigure}[b]{0.42\textwidth}
        \centering
        \includegraphics[scale=0.7]{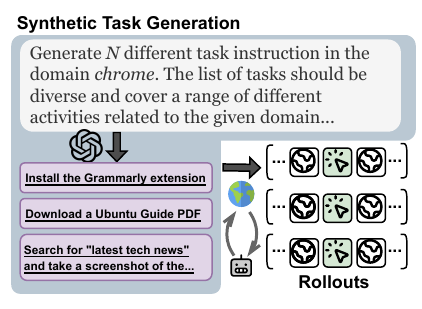}
        \caption{Generating synthetic tasks for rollouts.}
        \label{fig:synth_task_gen}
    \end{subfigure}
    \begin{subfigure}[b]{0.53\textwidth}
        \centering
        \includegraphics[scale=0.4]{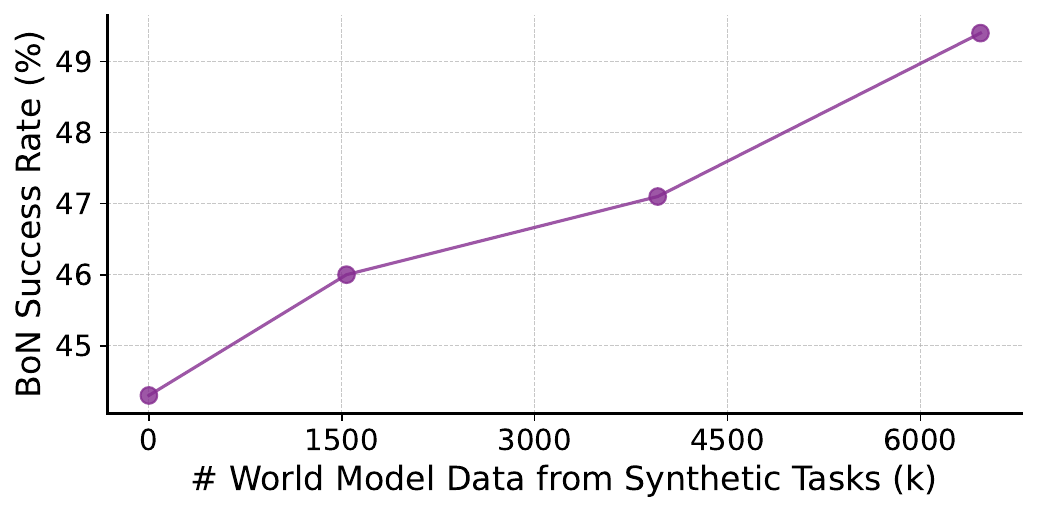}
        \caption{Test performance.}
        \label{fig:scaling_wm_results}
    \end{subfigure}
\caption{Scaling world model learning with synthetic tasks generated by GPT-4o. We use \judgeprefinetuneshort{} and train from our best model in \Cref{tbl:enable_v1}. After world model training, we perform one-round of policy training using the same set of policy learning data.}
\label{fig:scaling_wm_fig}
\vspace{-4mm}
\end{figure}

%% file: floats/iterative_training_v1fig.tex
\begin{figure}[t!]
    \begin{subfigure}[b]{0.4\textwidth}
        \centering
        \includegraphics[scale=0.35]{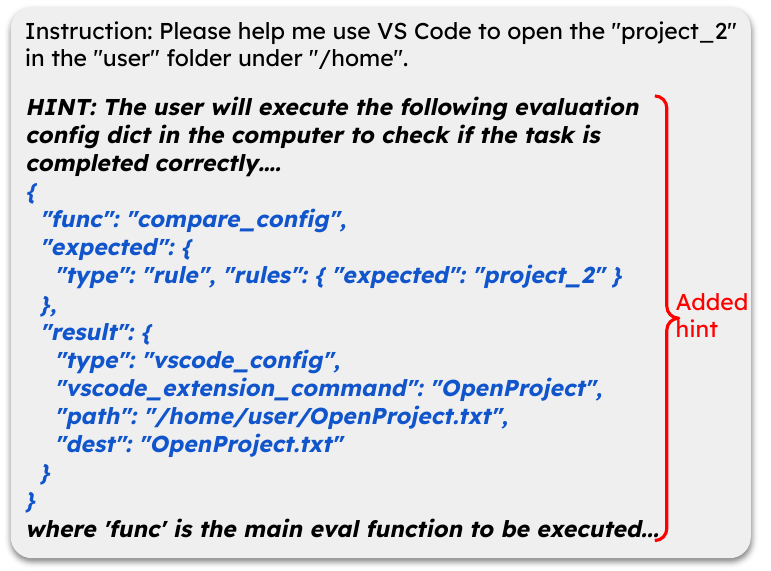}
        \caption{Rationalization with evaluation hint.}
        \label{fig:rationalization_expl}
    \end{subfigure}
    \begin{subfigure}[b]{0.53\textwidth}
        \centering
        \includegraphics[scale=0.35]{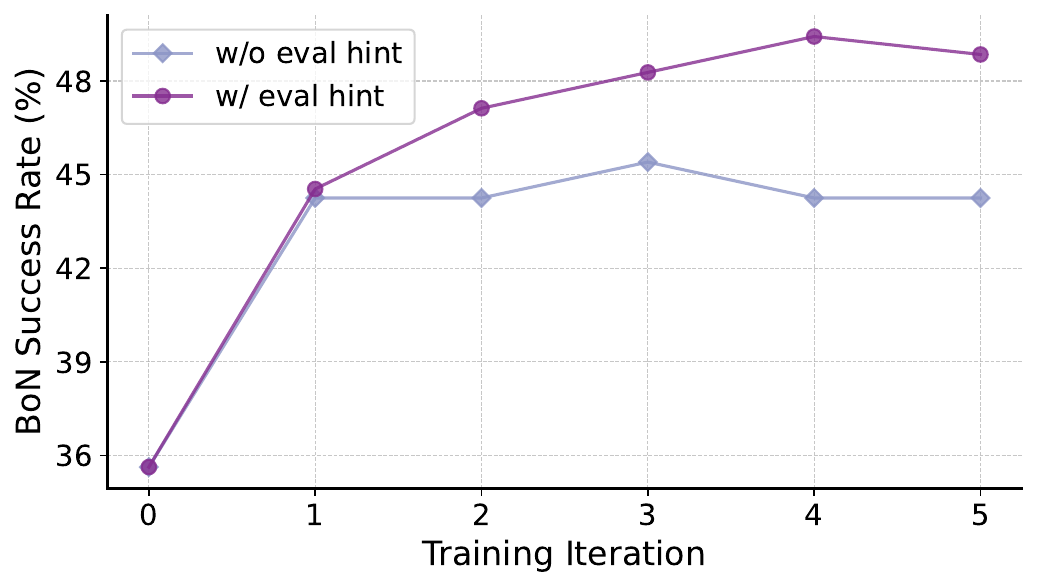}
        \caption{Test performance.}
        \label{fig:iterative_policy_results}
    \end{subfigure}
\caption{Iterative policy learning with and without adding evaluation configuration as hints. The added evaluation dictionary (in \textcolor{blue}{blue}) is part of the task configuration provided by OSWorld.}
\label{fig:iterative_policy_training}
\vspace{-5mm}
\end{figure}


%% file: floats/wm_acc_v1.tex
\begin{wraptable}[8]{r}{0.4\textwidth}
  \vspace{-15.4mm}
  \centering
  \caption{
  Measuring the world model accuracy (\emph{Acc}) and its correlation (\emph{$r$}) with task success rate.
  }
  \label{tbl:wm_acc_v1.1}
  \scalebox{0.74}{
    \begin{tabular}{l c l ccc cccc}
      \toprule
      \multirow{2}{*}{Method} & 
      \multicolumn{2}{c}{Policy} &
      \multicolumn{2}{c}{World Model}
      \\
      \cmidrule(lr){2-3}
      \cmidrule(lr){4-5}
      & \multicolumn{1}{c}{Avg} & 
      \multicolumn{1}{c}{BoN} &
      \multicolumn{1}{c}{Acc} &
      \multicolumn{1}{c}{$r$}\\
      \midrule
      R1-distill-70B
      & 7.6 & 15.5
      & 31.3 & 0.32
      \\
      R1
      & 31.2 & 44.8
      & 53.1 & 0.37
      \\
      \midrule
      \wmreshort{}(R1)
      & 22.6 & 35.6
      & 38.9 & 0.45
      \\
      \quad +RFT
      & 23.2 & 38.5
      & 46.6 & 0.37
      \\
      \quad +\judgeprefinetuneshort{}
      & 26.6 & 44.3
      & 55.7 & 0.44
      \\
      \bottomrule
    \end{tabular}
    }
\end{wraptable}

%% file: floats/minimal_text_example.tex
\begin{wrapfigure}[13]{R}{0.4\textwidth}
    \vspace{-10mm}
    \centering
    \includegraphics[scale=0.37]{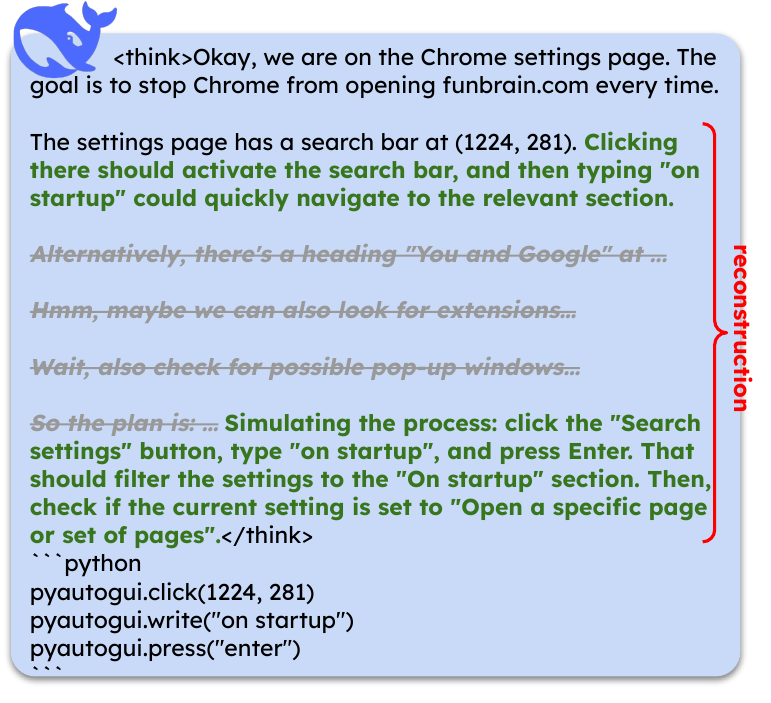}
    \caption{Example reconstruction in \wmreshort{}. World simulation is in \textcolor{teal}{green}.}
    \label{fig:minimal_example}
\end{wrapfigure}

%% file: floats/wm_judge_example.tex
\begin{wrapfigure}[24]{R}{0.52\textwidth}
    \vspace{-14mm}
    \centering
    \includegraphics[scale=0.55]{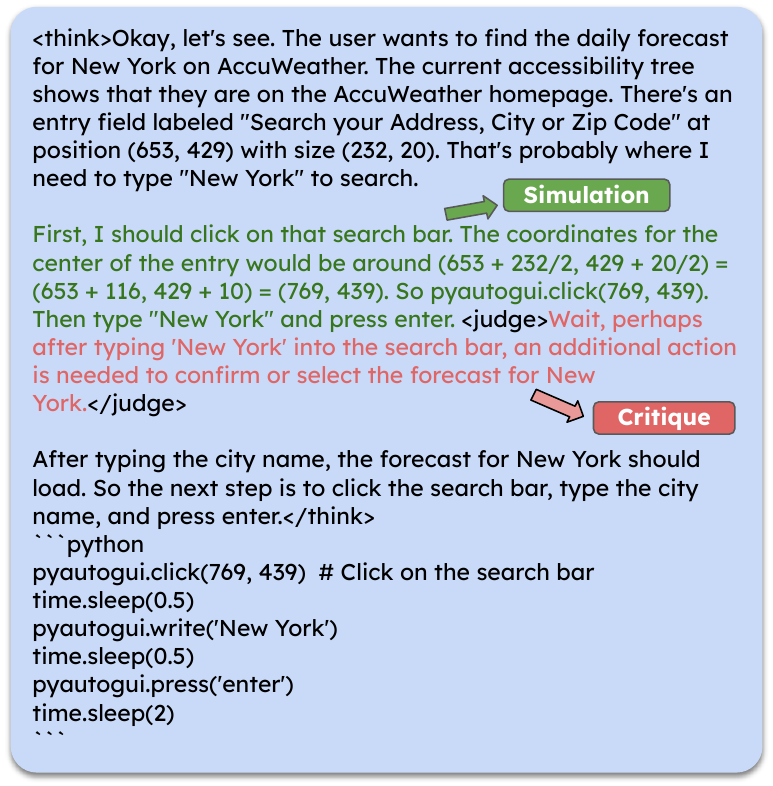}
    \caption{Example simulation critique data. 
    We first prompt GPT-4o to extract the world model simulation (shown in \textcolor{teal}{green}) corresponding to the final action; and then prompt GPT-4o to generate an critique (shown in \textcolor{red}{red}) based on the extracted simulation and the next state (see \Cref{tbl:wm_judge_prompt}).
    During training, all tokens except for text in \textcolor{red}{red} is masked.
    }
  
    \label{fig:judge_data_example}
\end{wrapfigure}

%% file: floats/wm_training_output_example.tex
\begin{figure}[t!]
\centering
\includegraphics[width=0.9\linewidth]{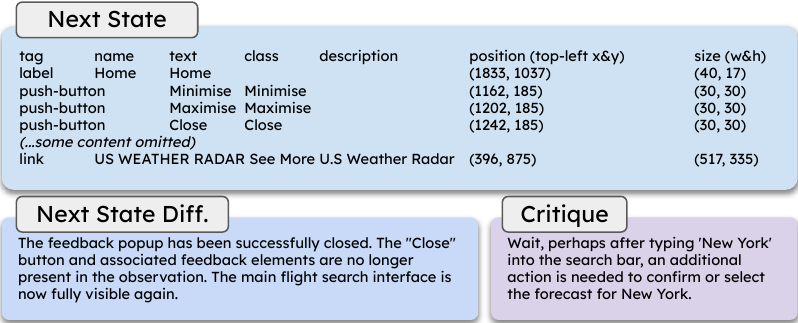}
\caption{Example next-state, state-difference, and simulation-critique data used in \internalDynatrainingshort{} training.}
\label{fig:wm_data_example}
\end{figure}

%% file: floats/train_test_set.tex
\begin{wraptable}[10]{R}{0.28\textwidth}
    \vspace{-10mm}
    \centering
    \caption{Number of tasks per domain in the training and test set.}
    \label{tbl:train_test_set}
    \begin{tabular}{l cc}
    \toprule
    Domain & Train & Test \\
    \midrule
    OS & 57 & 31\\
    VSCode & 46 & 38\\
    Chrome & 88 & 54\\
    GIMP & - & 32\\
    TBird & - & 19\\
    \bottomrule
    \end{tabular}
\end{wraptable}

%% file: floats/full_training_pipeline.tex
\begin{figure}[t!]
    \centering
    \includegraphics[scale=1.0]{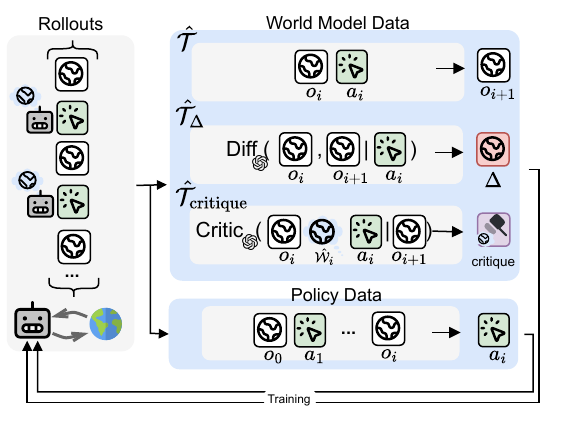}
    \caption{Illutration of \internalDynatraining{}. Given a set of rollout trajectories, we first perform world model training and then perform policy training.}
    \label{fig:full_training_pipeline}
\end{figure}

%% file: floats/full_domain_results.tex
\begin{table}[t!]
  \centering
  \caption{Model performance on additional domains (VLC, LibreOffice Impress, LibreOffice Writer, LibreOffice Calc, and Workflow). All model training are based on Qwen-2.5-32B-Instruct.
  We excluded these domains from \Cref{sec:Experiments} because they often require \emph{visual} interactions (\eg VLC involves video controls and LibreOffice Impress often needs slide edits) that the accessibility tree often cannot represent.
  }
  \scalebox{0.76}{
  \begin{tabular}{ll ccccc}
    \toprule
    \multirow{2}{*}{Method} &
    \multirow{2}{*}{Gen. Token} & 
    \multicolumn{5}{c}{Success Rate}
    \\
    \cmidrule(lr){3-7}
    & (10\%-90\%) &
    \multicolumn{1}{c}{VLC (17)} &
    \multicolumn{1}{c}{Impress (47)} &
    \multicolumn{1}{c}{Writer (23)} &
    \multicolumn{1}{c}{Calc (47)} &
    \multicolumn{1}{c}{Workflow (101)} 
    \\
    \midrule

    
    - (Qwen2.5-32B-Instruct) 
    & 0.2x-1.0x
    & 11.76 & 6.38 & 4.35 & 0.00 & 6.93
    \\


    \midrule
    \directdistillshort{}(R1) 
    & 2.0x-7.7x
    & 17.65 & 4.26 & 8.70 & 2.13 & 6.93
    \\
    \wmreshort{}(R1) 
    & 1.3x-2.9x
    & 11.76 & 2.12 & 13.04 & 2.13 & 7.92
    \\
    \bottomrule
\end{tabular}
  }
  \vspace{-10pt}
  \label{tbl:full_domain_results}
\end{table}

%% file: floats/test_aug_tasks.tex
\begin{table}
    \caption{Example augmented tasks. We modify the task instruction and/or the initialization configuration, such that action sequences that can correctly complete the original task does \emph{not} complete the augmented tasks.}
    \label{tbl:test_aug_task_examples}
    \begin{tabular}{p{1.52cm} p{11.51cm}}
        \toprule
        Domain & Task\\
        \midrule
        OS & 
        \textcolor{gray}{// original task}
        
        Can you remove the first favorite app from 'favorites'?"
        \\
        \cmidrule{2-2}
        & 
        \textcolor{gray}{// augmented tasks}

        Can you remove thunderbird and google chrome from the 'favorites' apps?

        Can you remove all the favorite apps except for thunderbird from 'favorites'?"
        \\
        \midrule
        VSCode
        &
        \textcolor{gray}{// original task}
        
        Please help me install the autoDocstring extension in VS Code."
        \\
        \cmidrule{2-2}
        & 
        \textcolor{gray}{// augmented tasks}

        Please help me install the Live Server extension by Ritwick Dey in VS Code.

        Please help me install the Auto Docstring and the Docker extension in VS Code.
        \\
        \midrule
        Chrome
        &
        \textcolor{gray}{// original task}

        Find a men's T-Shirt that is in large size with a stripe pattern, short sleeve and under the Sales\&Discount.
        \\
        \cmidrule{2-2}
        & 
        \textcolor{gray}{// augmented tasks}

        Find a men's T-Shirt that is blue, sleeveless and in XL size on Macy's website.

        Find a listing of woman's sandals with ankle strap on Macy's website."
        \\
        \bottomrule
    \end{tabular}
\end{table}

%% file: floats/wm_distill_prompt.tex
\begin{table}
    \caption{Prompts used abbreviate thinking in \wmreshort{}. Generated response is shown in \textcolor{blue}{blue}.
    }
    \label{tbl:wm_distill_prompt}
    \begin{tabular}{p{1.52cm} p{11.51cm}}
        \toprule
        Role & Prompt\\
        \midrule
        
        System & You are an agent which follow my instruction and perform desktop computer tasks as instructed (...some content omitted)

        You are asked to complete the following task: (...some content omitted)
        \\
        \cmidrule{2-2}
        User & 
        Given the info from accessibility tree as below:
        \newline
        
        tag	name	text	class	description	position (top-left x\&y)	size (w\&h)
        
label	Home	Home			(1833, 1037)	(40, 17)

(...some content omitted)

push-button	Trash	""			(0, 784)	(70, 64)

toggle-button	Show Applications	""			(0, 1010)	(70, 70)
\newline

        What's the next step that you will do to help with the task?
        \\
        \cmidrule{2-2}
        Assistant & 
        <think>Okay, the user is trying to (...some content omitted)</think>
        
        ```python
        
        pyautogui.click(x=35, y=1045)  \# Click Show Applications
        
        time.sleep(1)
        
        pyautogui.write(`terminal')
        
        (...some content omitted)
        
        ```
        \\
        \cmidrule{2-2}
        User &
        ACTIONS TO BE EXECUTED:

        ```python
        
        pyautogui.click(x=35, y=1045)  \# Click Show Applications
        
        time.sleep(1)
        
        pyautogui.write(`terminal')
        
        (...some content omitted)
        
        ```
        \newline
        
        Your task now is to ABBREVIATE the THINKING process (inside the <think> xxx </think> part) of the previous response.

        Specifically, you need to KEEP THE FOLLOWING INFORMATION in the thinking section, while removing others as much as possible:

        1. Thoughts related to verification: checking whether previous action has succeeded/failed. This is usually the first paragraph in the THINKING process.
        
        2. Thoughts related to what could be potential actions to do next, if mentioned. This usually follows after the verification process, (...some content omitted)
        
        3. Thoughts related to final action simulation: simulating the outcome/process of the 'ACTIONS TO BE EXECUTED' section (...some content omitted)
        
        4. Other thoughts that you believe is necessary to logically connect step 2 to step 3 above.
        
        5. Implementation details such as code snippets related to part 2 and 3 above should also be KEPT.
        \newline
        
        $\{\{$one shot example$\}\}$
        \newline

        Now, abbreviate the thinking process in the <think> xxx </think> section of the previous response. Your output should:

        - keep the **ORIGINAL FORMATTING AND TONE** of the response (...some content omitted)

        - only **REMOVE UNNECESSARY THINKING PARTS WIHOUT TOUCHING ANYTHING ELSE**. You can ONLY rephrase a few sentences if they are necessary (...some content omitted)

        - logically **CONNECT to the 'ACTIONS TO BE EXECUTED' section**(...some content omitted)

        Respond in the following format. DO NOT generate anything else.
        Abbreviated thinking process: <think>Okay, ...(your abbreviated thinking process)...</think>
        \\
        \cmidrule{2-2}
        Assistant & \textcolor{blue}{<think>Okay, the user is trying to (... some content omitted)</think>}\\
        \bottomrule
    \end{tabular}
\end{table}

%% file: floats/diff_n_judge_prompt.tex
\begin{table}
    \caption{Prompts used to obtain $\Delta(o_i,o_{i+1})$ data. Generated response is highlighted in \textcolor{blue}{blue}.
    }
    \label{tbl:wm_diff_prompt}
    \begin{tabular}{p{1.5cm} p{11.5cm}}
        \toprule
        Role & Prompt\\
        \midrule
        
        System & You are an agent which follow my instruction and perform desktop computer tasks as instructed (...some content omitted)

        You are asked to complete the following task: (...some content omitted)
        \\
        \cmidrule{2-2}
        User & 
        Given the info from accessibility tree as below:
        \newline
        
        tag	name	text	class	description	position (top-left x\&y)	size (w\&h)
        
label	Home	Home			(1833, 1037)	(40, 17)

(...some content omitted)

toggle-button	Show Applications	""			(0, 1010)	(70, 70)
\newline

        What's the next step that you will do to help with the task?
        \\
        \cmidrule{2-2}
        Assistant & 
        <think>Okay, the user is trying to (...some content omitted)</think>
        
        ```python
        
        pyautogui.click(x=35, y=1045)  \# Click Show Applications
        
        time.sleep(1)
        
        pyautogui.write(`terminal')
        
        (...some content omitted)
        
        ```
        \\
        \cmidrule{2-2}
        User &
        ACTIONS TO BE EXECUTED:

        ```python
        
        pyautogui.click(x=35, y=1045)  \# Click Show Applications
        
        time.sleep(1)
        
        pyautogui.write(`terminal')
        
        (...some content omitted)
        
        ```
        \newline

        NEXT OBSERVATION AFTER EXECUTION:
        
        tag	name	text	class	description	position (top-left x\&y)	size (w\&h)
        
label	Home	Home			(1833, 1037)	(40, 17)

(...some content omitted)

toggle-button	Show Applications	""			(0, 1010)	(70, 70)
\newline

Your task now is to describe changes in the next observation caused by the previous action. Specifically, you need to describe what changes, if any, occured on the 'NEXT OBSERVATION AFTER EXECUTION' after executing the 'ACTIONS TO BE EXECUTED'. The description should:

1. give high-level and functional description in detail after the action.

2. focus solely on describing the changes that are caused by the action.

3. if the 'NEXT OBSERVATION AFTER EXECUTION' is identical to the previous observation, say 'no changes'.

4. keep your description concise. **DO NOT explain why** certain changes ocurred, only describe the changes in observation.
\newline

Now, describe the changes in 'NEXT OBSERVATION AFTER EXECUTION' that are likely caused by 'ACTIONS TO BE EXECUTED' section. Respond in the following format. DO NOT generate anything else. Keep your answer within 3 sentences.

Changes description: <your description of the changes>
        \\
        \cmidrule{2-2}
        Assistant & \textcolor{blue}{Changes description: The terminal has been (...some content omitted)}\\
        \bottomrule
    \end{tabular}
\end{table}

\begin{table}
    \caption{Prompts used to obtain $\textrm{Critic}(o_i,a_i|o_{i+1})$ data. Generated response is highlighted in \textcolor{blue}{blue}.
    Extracted world model simulation from the agent's response is shown in \textcolor{teal}{green}.
    }
    \label{tbl:wm_judge_prompt}
    \begin{tabular}{p{1.5cm} p{11.5cm}}
        \toprule
        Role & Prompt\\
        \midrule
        
        System & You are an agent which follow my instruction and perform desktop computer tasks as instructed (...some content omitted)

        You are asked to complete the following task: (...some content omitted)
        \\
        \cmidrule{2-2}
        User & 
        Given the info from accessibility tree as below:
        \newline
        
        tag	name	text	class	description	position (top-left x\&y)	size (w\&h)
        
label	Home	Home			(1833, 1037)	(40, 17)

(...some content omitted)

toggle-button	Show Applications	""			(0, 1010)	(70, 70)
\newline

        What's the next step that you will do to help with the task?
        \\
        \cmidrule{2-2}
        Assistant & 
        <think>Okay, the user is trying to (...some content omitted)</think>
        
        ```python
        
        pyautogui.click(x=35, y=1045)  \# Click Show Applications
        
        time.sleep(1)
        
        pyautogui.write(`terminal')
        
        (...some content omitted)
        
        ```
        \\
        \cmidrule{2-2}
        User &
        ACTIONS TO BE EXECUTED:

        ```python
        
        pyautogui.click(x=35, y=1045)  \# Click Show Applications
        
        time.sleep(1)
        
        pyautogui.write(`terminal')
        
        (...some content omitted)
        
        ```
        \newline

        RELATED THINKING FROM PREVIOUS RESPONSE:

        \textcolor{teal}{Clicking on the `Show Applications' will open an application grid. Then, we can (...some content omitted).}
        \newline

        NEXT OBSERVATION AFTER EXECUTION:
        
        tag	name	text	class	description	position (top-left x\&y)	size (w\&h)
        
label	Home	Home			(1833, 1037)	(40, 17)

(...some content omitted)

toggle-button	Show Applications	""			(0, 1010)	(70, 70)
\newline

Your task now is to JUDGE if the action (in the 'ACTIONS TO BE EXECUTED' section) achieved the intended effect in the 'ACTUAL NEXT STATE' section. Specifically, you need to:

1. decide whether the 'RELATED THINKING FROM PREVIOUS RESPONSE' section is correct based on the 'ACTUAL NEXT STATE' section.

2. provide a confidence score from 0 to 5 for the above judgment, where 0 means 'not at all confident' and 5 means 'very confident'.

3. if you think 'RELATED THINKING FROM PREVIOUS RESPONSE' is incorrect, provide a one-sentence reason (...some content omitted)

4. if you thin the 'RELATED THINKING FROM PREVIOUS RESPONSE' is correct, provide a one-sentence reason (...some content omitted)

NOTE: if the thinking is CORRECT, then the hint should:

- (...some content omitted)
- start with 'Yes, ...'.

NOTE: if the thinking is INCORRECT, then the hint should:

- (...some content omitted)
- start with 'Wait, ...'. (...some content omitted)
        \\
        \cmidrule{2-2}
        Assistant & \textcolor{blue}{
        is\_simulation\_correct: False
        \newline
        confidence: 5
        \newline
        reason: Terminal application is not present in the next state (...some content omitted)
        \newline
        in\_context\_hint: Wait, maybe we need to wait a bit before typing `terminal', or ...}
        \\
        \bottomrule
    \end{tabular}
\end{table}

%% file: floats/task_gen_prompt.tex
\begin{table}
    \caption{Prompts used to generate synthetic task instructions in domains excluding GIMP. Generated response is highlighted in \textcolor{blue}{blue}, with \{\textit{domain}\} set to chrome.
    }
    \label{tbl:task_generation_prompt}
    \begin{tabular}{p{1.5cm} p{11.5cm}}
        \toprule
        Role & Prompt\\
        \midrule
         User & You are a helpful assistant. Imagine you are in an Ubuntu VMware environment, with access to applications like os, vscode, chrome, gimp, thunderbird, etc.

Your mission is to generate a list of\textbf{ 10} task instructions that can be performed in this environment, and the tasks should be related to a given domain (application), which means the tasks can be solved mainly by this application.

You should guarantee:

(1) The tasks should be clear to understand and should be related to common activities that a user might perform in a virtual machine environment.

(2) The tasks should not be exploratory, i.e. whether the tasks are completed correctly should be verifiable.
You should avoid using words like "set a secure password", "install a specific software", etc. Instead, you should explicitly nominate these by making up them or using your knowledge.

(3) The tasks should be interesting and challenging, but not too complicated, i.e. a user may accomplish them in a few minutes.

(4) You don't need to consider the exact environment, such as the file structure and the application versions, etc. You can add details to the tasks as long as they look reasonable and not too complicated.

(5) The list of tasks should be diverse and cover a range of different activities related to the given domain.

(6) You should NOT add indexes or titles to the tasks. Directly list the task instructions split by new lines.

The domain for the tasks is ""\{\textit{domain}\}"".\\
          \cmidrule{2-2}
         Assistant & \textcolor{blue}{Access the Chrome settings and enable "Do Not Track" requests for all websites.}
         \newline

        \textcolor{blue}{Open Chrome and go to the YouTube website. Search for a video on sports topic and add it to the "Watch Later" playlist to ensure you can access it later.}
        \newline

        \textcolor{blue}{Set up Chrome to automatically translate web pages written in a foreign language to English using the built-in translation feature.}
        \newline

        \textcolor{blue}{Use Chrome to download a PDF version of a popular Ubuntu guide, ensuring it's saved to the "Downloads" folder.}
        \newline

        \textcolor{blue}{Use the "Save as PDF" feature in Chrome to save an online recipe from a cooking website to your desktop.}
        \newline

        \textcolor{blue}{Enable the reading list feature in Chrome's settings. Add three different articles from any technology news website to the reading list and mark one as read.}
        \newline

        \textcolor{blue}{Open Chrome and navigate to a tech news website like TechCrunch. Bookmark the homepage and create a new folder named "Tech News" in your bookmarks bar for future reference.}
        \newline

        \textcolor{blue}{Open Chrome and navigate to a weather website like weather.com. Search for the weather forecast of your city for the next week and take a screenshot of the results.}
        \newline

        \textcolor{blue}{Access Chrome's "Task Manager" by clicking "More tools" and "Task Manager." Identify the process consuming the most memory and end the task to free up resources.}
        \newline

        \textcolor{blue}{Open Chrome and navigate to the Chrome Web Store. Install the Grammarly extension and verify that it is enabled by checking the extensions list in the browser settings.}
        \\
         \bottomrule
    \end{tabular}
\end{table}

\begin{table}
    \caption{Prompt used to generate synthetic tasks for the GIMP domain.}
    \label{tbl:gimp_generation_prompt}
    \begin{tabular}{p{13cm}}
        \toprule
        GIMP Prompt\\
        \midrule
          You are a helpful assistant. Imagine you are in an Ubuntu VMware environment, with access to applications like os, vscode, chrome, gimp, thunderbird, etc. 
          
The followings are an example of a pair of task instruction and corresponding environment setup configuration for GIMP:

[Environment setup configuration]: $\{\{$config$\}\}$

[Task instruction]: $\{\{$instruction$\}\}$

Your mission is to generate a list of 10 task instructions which are also related to gimp and can all be performed with the SAME environment setup configuration as the above example.

You should guarantee: The tasks should be clear to understand and should be related to realistic activities. The tasks could be interesting, challenging, and properly complicated, i.e. a user may accomplish them in a few minutes.

If the example's configuration is empty, you should ONLY consider tasks that does not require any image operations, such as starting from a blank canvas, or changing the settings of gimp itself.

For the result you generate, please make sure: 

The list of tasks should be diverse and cover a range of different activities related to gimp. They should be DISTINCT from each other and the examples provided.

You should NOT add indexes or titles to your answers. Directly list the task instructions split by new lines.
        \\
         \bottomrule
    \end{tabular}
\end{table}

%% file: floats/wm_acc_prompt.tex
\begin{table}
    \caption{Prompts used to extract world model simulation. Generated response is in \textcolor{blue}{blue}.
    }
    \label{tbl:wm_acc_simextract_prompt}
    \begin{tabular}{p{1.5cm} p{11.5cm}}
        \toprule
        Role & Prompt\\
        \midrule
        
        System & You are an agent which follow my instruction and perform desktop computer tasks as instructed (...some content omitted)

        You are asked to complete the following task: (...some content omitted)
        \\
        \cmidrule{2-2}
        User & 
        Given the info from accessibility tree as below:
        \newline
        
        tag	name	text	class	description	position (top-left x\&y)	size (w\&h)
        
label	Home	Home			(1833, 1037)	(40, 17)

(...some content omitted)

toggle-button	Show Applications	""			(0, 1010)	(70, 70)
\newline

        What's the next step that you will do to help with the task?
        \\
        \cmidrule{2-2}
        Assistant & 
        <think>Okay, the user is trying to (...some content omitted)</think>
        
        ```python
        
        pyautogui.click(x=35, y=1045)  \# Click Show Applications
        
        time.sleep(1)
        
        pyautogui.write(`terminal')
        
        (...some content omitted)
        
        ```
        \\
        \cmidrule{2-2}
        User &
        ACTIONS TO BE EXECUTED:

        ```python
        
        pyautogui.click(x=35, y=1045)  \# Click Show Applications
        
        time.sleep(1)
        
        pyautogui.write(`terminal')
        
        (...some content omitted)
        
        ```
        \newline

        Your task now is to ANNOTATE the thinking process of the previous action. Specifically, you need to EXTRACT the simulation process behind the 'ACTIONS TO BE EXECUTED' section above. You should find them inside the previous response (e.g., inside <think> xxx </think>).

        The extracted simulation process:

        1. SHOULD mention the high level plan of these actions, if exists

        2. SHOULD mention the ALL the action in the 'ACTIONS TO BE EXECUTED' section AND the effect each action was supposed to have. If this is missing for some actions, you can fill in your best guess based on the context.

        3. SHOULD ONLY mention any action or effect included in the 'ACTIONS TO BE EXECUTED' section.

        4. SHOULD keep the original formatting and tone of the response.

        5. If you DID NOT FIND ANY such simluation process exists in the previous response, you should still follow rule 2 but set the 'has\_world\_model\_simulation' to False.
        \newline

        FOR EXAMPLE, given the following action and original response:

        $\{\{$one shot example$\}\}$
        \newline

        Now, extract the simulation process for the 'ACTIONS TO BE EXECUTED' section.
        \\
        \cmidrule{2-2}
        Assistant & \textcolor{blue}{
        has\_world\_model\_simulation: True
        \newline
        simulation\_process: Clicking `Show Applications' should open an application grid, with search bar at... In the search bar we can type (...some content omitted)
        }
        \\
        \bottomrule
    \end{tabular}
\end{table}
\begin{table}
    \caption{Prompts used to test simulation accuracy. World model simulation is in \textcolor{teal}{green}. Generated response is in \textcolor{blue}{blue}.
    }
    \label{tbl:wm_acc_test_prompt}
    \begin{tabular}{p{1.5cm} p{11.5cm}}
        \toprule
        Role & Prompt\\
        \midrule
        
        System & You are an agent which follow my instruction and perform desktop computer tasks as instructed (...some content omitted)

        You are asked to complete the following task: (...some content omitted)
        \\
        \cmidrule{2-2}
        User & 
        Given the info from accessibility tree as below:
        \newline
        
        tag	name	text	class	description	position (top-left x\&y)	size (w\&h)
        
label	Home	Home			(1833, 1037)	(40, 17)

(...some content omitted)

toggle-button	Show Applications	""			(0, 1010)	(70, 70)
\newline

        What's the next step that you will do to help with the task?
        \\
        \cmidrule{2-2}
        Assistant & 
        <think>Okay, the user is trying to (...some content omitted)</think>
        
        ```python
        
        pyautogui.click(x=35, y=1045)  \# Click Show Applications
        
        time.sleep(1)
        
        pyautogui.write(`terminal')
        
        (...some content omitted)
        
        ```
        \\
        \cmidrule{2-2}
        User &
        RELATED THINKING FROM PREVIOUS RESPONSE:

        \textcolor{teal}{Clicking on the `Show Applications' will open an application grid. Then, we can (...some content omitted).}
        \newline
        
        ACTIONS TO BE EXECUTED:

        ```python
        
        pyautogui.click(x=35, y=1045)  \# Click Show Applications
        
        time.sleep(1)
        
        pyautogui.write(`terminal')
        
        (...some content omitted)
        
        ```
        \newline

        ACTUAL NEXT STATE:
        
        tag	name	text	class	description	position (top-left x\&y)	size (w\&h)
        
label	Home	Home			(1833, 1037)	(40, 17)

(...some content omitted)

toggle-button	Show Applications	""			(0, 1010)	(70, 70)
\newline

Your task now is to JUDGE if the action (in the 'ACTIONS TO BE EXECUTED' section) achieved the intended effect in the 'ACTUAL NEXT STATE' section. Specifically, you need to:

1. decide whether the 'RELATED THINKING FROM PREVIOUS RESPONSE' section is correct based on the 'ACTUAL NEXT STATE' section.

2. provide a confidence score from 0 to 5 for the above judgment, where 0 means 'not at all confident' and 5 means 'very confident'.

3. if you think the 'RELATED THINKING FROM PREVIOUS RESPONSE' is incorrect, provide a one-sentence reason of which part of the thinking may cause the error.

4. if you think the 'RELATED THINKING FROM PREVIOUS RESPONSE' is correct, provide a one-sentence reason why you think so.
        \\
        \cmidrule{2-2}
        Assistant & \textcolor{blue}{
        is\_simulation\_correct: False
        \newline
        confidence: 5
        \newline
        reason: Terminal application is not present in the next state (...some content omitted)}
        \\
        \bottomrule
    \end{tabular}
\end{table}